\documentclass[11pt, a4paper, logo, copyright]{googledeepmind}

\usepackage[authoryear, sort&compress, round]{natbib}

\usepackage{cleveref}
\usepackage{url}
\usepackage{wrapfig,booktabs}
\usepackage{xcolor}         
\usepackage{tikz}
\usetikzlibrary{patterns}
\usepackage{colortbl}
\usepackage{arydshln}
\usepackage{adjustbox}
\usepackage{multirow}
\usepackage{enumerate}
\usepackage{enumitem}
\usepackage{subfigure}
\usepackage{gensymb}
\usepackage{xspace} 
\usepackage{subcaption} 
\usepackage{graphicx}
\usepackage{makecell}  
\usepackage{minitoc}             
\usepackage{makeidx}
\usepackage{booktabs}
\usepackage{bbm}
\usepackage{enumerate}
\usepackage{algorithm}
\usepackage{algorithmic}
\usepackage{colortbl}
\usepackage{psfrag}
\usepackage{adjustbox}
\usepackage{wrapfig}
\usepackage{xspace}
\usepackage{amssymb}
\usepackage{multirow}
\usepackage{afterpage}
\usepackage{float}
\usepackage[toc,page,header]{appendix}
\usepackage{minitoc}
\usepackage{placeins} 
\usepackage{titletoc} 

\colorlet{darkgreen}{green!65!black}
\colorlet{darkblue}{blue!75!black}
\colorlet{darkred}{red!80!black}
\definecolor{lightblue}{HTML}{0071bc}
\definecolor{lightgreen}{HTML}{39b54a}
\definecolor{manyshot}{HTML}{6969ff}
\definecolor{medshot}{HTML}{f7c600}
\definecolor{fewshot}{HTML}{ff6969}
\definecolor{mypurple}{HTML}{412F8A}
\definecolor{myorange}{HTML}{fc8e62}

\definecolor{deemph}{gray}{0.55}

\definecolor{textgreen}{RGB}{57, 172, 57}
\definecolor{textred}{RGB}{200, 10, 10}
\definecolor{textgray}{RGB}{100, 100, 100}
\definecolor{visiongold}{RGB}{230, 184, 0}
\definecolor{speechpurple}{RGB}{204, 0, 255}
\definecolor{dataprep}{RGB}{38, 189, 128}
\definecolor{modeltraining}{RGB}{38, 189, 128}
\definecolor{backgroundcol}{RGB}{232, 230, 230}
\definecolor{gold}{rgb}{225, 215, 200} 
\definecolor{navyblue}{RGB}{40, 66, 200} 
\definecolor{orange}{RGB}{255,127,80} 
\definecolor{pink}{RGB}{219,112,147} 
\newcommand{\grayrow}{\rowcolor[gray]{.9}}
\definecolor{baselinecolor}{gray}{.95}

\newcommand{\AccelerometerCircle}[1][0.9]{%
    \tikz[baseline=(char.base)]\node[shape=rectangle,draw=black,inner sep=2pt,line width=0.5pt,fill=textgreen,text=white,scale=#1] (char) {ACC};\xspace
}
\newcommand{\PPGCircle}[1][0.9]{%
    \tikz[baseline=(char.base)]\node[shape=rectangle,draw=black,inner sep=2pt,line width=0.5pt,fill=visiongold,text=white,scale=#1] (char) {PPG};\xspace
}
\newcommand{\EDACircle}[1][0.92]{%
    \tikz[baseline=(char.base)]\node[shape=rectangle,draw=black,inner sep=2pt,line width=0.5pt,fill=lightblue,text=white,scale=#1] (char) {SCL};\xspace
}

\newcommand{\ECGCircle}[1][0.92]{%
    \tikz[baseline=(char.base)]\node[shape=rectangle,draw=black,inner sep=1pt,line width=0.5pt,fill=textred,text=white,scale=#1] (char) {ECG};\xspace
}

\newcommand{\TempCircle}[1][0.92]{%
    \tikz[baseline=(char.base)]\node[shape=rectangle,draw=black,inner sep=1pt,line width=0.5pt,fill=orange,text=white,scale=#1] (char) {TMP};\xspace
}
\newcommand{\AltCircle}[1][0.92]{%
    \tikz[baseline=(char.base)]\node[shape=rectangle,draw=black,inner sep=1pt,line width=0.5pt,fill=pink,text=white,scale=#1] (char) {ALT};\xspace
}

\newcommand{\cmark}{\textcolor{textgreen}{\ding{51}}}%
\newcommand{\xmark}{\textcolor{textred}{\ding{55}}}%
\usepackage{pifont}

\usepackage{makecell, tabularx}
\newcolumntype{L}{>{\RaggedRight}X}
\usepackage{siunitx}

\title{Scaling Wearable Foundation Models}

\correspondingauthor{$\{$xliucs, dmcduff$\}$@google.com}

\reportnumber{} 

\renewcommand{\today}{2024-10-03}

\author[$\circ$,1]{Girish Narayanswamy}
\author[$\circ$,$\dagger$,1]{Xin Liu}
\author[1]{Kumar Ayush}
\author[1]{Yuzhe Yang}
\author[1]{Xuhai Xu}
\author[1]{Shun Liao}
\author[1]{Jake Garrison}
\author[1]{Shyam Tailor}
\author[1]{Jake Sunshine}
\author[1]{Yun Liu}
\author[1]{Tim Althoff}
\author[1]{Shrikanth Narayanan}
\author[2]{Pushmeet Kohli}
\author[1]{Jiening Zhan}
\author[1]{Mark Malhotra}
\author[1]{Shwetak Patel}
\author[1]{Samy Abdel-Ghaffar}
\author[$\dagger$,1]{Daniel McDuff}

\affil[$\circ$]{Co-first}
\affil[$\dagger$]{Corresponding Author}
\affil[1]{Google Research}
\affil[2]{Google DeepMind}

\newcommand{\lsm}{\texttt{LSM}\xspace}

\begin{abstract}
Wearable sensors have become ubiquitous thanks to a variety of health tracking features. The resulting continuous and longitudinal measurements from everyday life generate large volumes of data; however, making sense of these observations for scientific and actionable insights is non-trivial. Inspired by the empirical success of generative modeling, where large neural networks learn powerful representations from vast amounts of text, image, video, or audio data, we investigate the scaling properties of sensor foundation models across compute, data, and model size. Using a dataset of up to 40 million hours of in-situ heart rate, heart rate variability, electrodermal activity, accelerometer, skin temperature, and altimeter per-minute data from over 165,000 people, we create \lsm, a multimodal foundation model built on the largest wearable-signals dataset with the most extensive range of sensor modalities to date. Our results establish the scaling laws of \lsm for tasks such as imputation, interpolation and extrapolation, both across time and sensor modalities. Moreover, we highlight how \lsm enables sample-efficient downstream learning for tasks like exercise and activity recognition.
\vspace{-0.1cm}
\end{abstract}

\begin{document}

\maketitle

\newenvironment{Itemize}{
    \begin{itemize}[leftmargin=*]
    \setlength{\itemsep}{0pt}
    \setlength{\topsep}{0pt}
    \setlength{\partopsep}{0pt}
    \setlength{\parskip}{0pt}}
{\end{itemize}}
\setlength{\leftmargini}{9pt}

\section{Introduction}
\vspace{-0.1cm}

Wearable devices that monitor physiological and behavioral signals have become ubiquitous. Increasing evidence suggests that these devices can significantly contribute to promoting healthy behaviors \citep{ringeval2020fitbit}, detecting diseases \citep{yang2022artificial}, and enhancing the design and implementation of treatments \citep{munos2016mobile}. These devices generate large volumes of continuous, longitudinal, and multimodal data. However, raw data from sensors such as accelerometers or photoplethysmography (PPG) hardware are often challenging for both consumers and experts to interpret. To address this issue, algorithms have been developed to translate sensor outputs into more meaningful representations, such as step counts and heart rate.

Historically, algorithms for wearable sensors have relied on supervised, discriminative models designed to detect specific events or activities \citep{lubitz2022detection}. This approach, however, faces several significant limitations. First, the \textit{limited volume and severe data imbalance} of labeled events results in large amounts of valuable \emph{unlabeled} data being left unused. Second, supervised models are typically trained for \textit{a single task} (e.g., classification), producing representations that may not generalize well to other tasks. Third, training data is often collected from \textit{small study populations} (usually involving only tens or hundreds of participants), leading to a lack of diversity in the data.

\begin{figure*}[h]
    \centering
    \includegraphics[width=\textwidth]{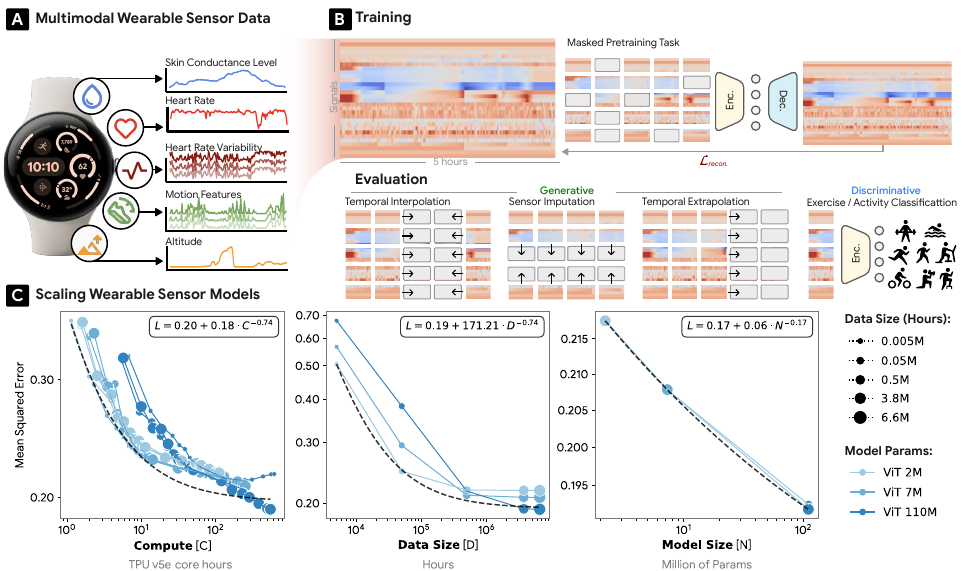}
    \vspace{-16pt}
    \caption{\textbf{Scaling foundation models on wearable data.} Making sense of physiological and behavioral signals derived from wearables is challenging. \textbf{(A)} We present a systematic scaling analysis of sensor models using up to 40 million hours of multimodal data from over 165,000 people. \textbf{(B)} Using a random masking pretext task, we evaluate on tasks of imputation, forecasting, and downstream classification. \textbf{(C)} Experiments show scaling compute, data, and model size are all effective. Scaling is shown on the random imputation task.}
    \label{fig:teaser} 
    \vspace{-0.3cm}
\end{figure*}

Self-supervised learning (SSL) using generic pretext tasks \citep{noroozi2017representation,caron2018deep,yangsimper} can yield versatile representations that are useful for a wide range of downstream applications.
SSL allows for the use of a much larger proportion of available data without being restricted to labeled data regions (e.g., a limited number of subjects who self-report labels for exercises/activities). 
These advantages have motivated efforts to apply similar training strategies to build models from large volumes of unlabeled wearable data \citep{adaimi2024advancing,thapa2024sleepfm,yuan2024self,abbaspourazad2023large} (see Table~\ref{tab:related_work} for a summary).

Building on this, the empirical and theoretical success of scaling laws in neural models~\citep{kaplan2020scaling, bahri2024explaining} suggests that model performance improves predictably as compute, data, and model parameters increase. These findings raise a critical research question: \textbf{Do scaling laws apply to models trained on wearable sensor data?} We aim to investigate whether the principles that drive the scaling of neural networks in domains like language and vision also extend to large-scale, multimodal wearable sensor data. Understanding how scaling manifests in this context could not only shape model design but also enhance generalization across diverse tasks and datasets.

In this paper, we present the results of our scaling experiments on the largest and the most diverse wearable dataset published to date, comprising 40 million hours of multimodal sensor data from over 165,000 users (Fig.~\ref{fig:teaser}). Leveraging these data, we train a foundation model, referred to as the \textbf{Large Sensor Model} (\lsm), which is designed to capture generalizable representations across diverse populations, wearable sensor modalities, and downstream tasks.
We demonstrate the scaling properties of \lsm with respect to compute, data size, and model parameters, leading to substantial performance gains on generative imputation, interpolation and extrapolation as well as downstream discriminative tasks. Our contributions can be summarized as follows:
\begin{Itemize}
    \item Implementation of the largest study to date on the scaling behavior of sensor foundation models, encompassing 40 millions hours, over 165,000 users and multiple sensor modalities, including accelerometer, photoplethysmography (PPG), electrodermal activity (EDA), skin temperature, and altimeter signals.
    \item Identification of key strategies for training large-scale sensor foundation models (\lsm), and the \lsm's scaling properties with respect to compute, data size, and model parameters.
    \item Demonstration of the model's ability to impute, interpolate, and extrapolate across temporal and sensor modalities, with a particular focus on generalization to unseen users. 
    \item Verification that learned representations can be applied to downstream classification tasks, such as exercise and activity recognition, using ecologically valid, user-annotated events.
\end{Itemize}
 
\section{Related Work}

\textbf{Sensor Foundation Models.}
Recent advances have demonstrated improved accuracy, robustness, and generalizability of models for sensor data by utilizing self-supervised pretraining on large-scale corpora of behavioral and physiological signals \citep{yuan2024self,thapa2024sleepfm,merrill2023self}.
Existing sensor foundation models primarily leverage contrastive learning, creating positive and negative data pairs \citep{yuan2024self,thapa2024sleepfm,abbaspourazad2023large}. Yuan et al. (\citeyear{yuan2024self}) employ time domain augmentations (e.g., reversal, warping, permutation) to formulate the SSL task for motion data. Abbaspourazad et al. (\citeyear{abbaspourazad2023large}) adopt a similar strategy, incorporating Gaussian noise, time and magnitude warping, and channel swapping. Thapa et al. (\citeyear{thapa2024sleepfm}) generate data pairs using different sensory modalities. In contrast, we focus on \textit{masked input modeling} due to the generative capabilities that it offers and explore its properties when scaling compute, data size, and model size.  Compared to prior work we consider more sensor inputs, a larger data sample, and systematically investigate scaling laws (see Table~\ref{tab:related_work}). We also present contrastive baselines~\citep{assran2022masked,chen2020simple} where applicable.

\begin{table}[!t]
    \centering
    \caption{\textbf{Comparisons of studies on wearable sensor foundation models.}}
    \vspace{-5pt}
    \begin{tabular}{rcccccccccc}
    \toprule[1.5pt]
        \textbf{Study}   & \rotatebox{45}{\textbf{\# People}} & \rotatebox{45}{\textbf{\# Hours}}  & \multicolumn{6}{c}{\rotatebox{45}{\textbf{Sensors}}} & \rotatebox{45}{\textbf{Generative}} \\
        &  (000s) &  (000s) & \rotatebox{90}{\ECGCircle} & \rotatebox{90}{\PPGCircle} & \rotatebox{90}{\AccelerometerCircle}  & \rotatebox{90}{\EDACircle} & \rotatebox{90}{\TempCircle} & \rotatebox{90}{\AltCircle}  \\
        \midrule
        \citet{adaimi2024advancing} & 0.05 & 0.20  & \xmark & \xmark & \cmark & \xmark & \xmark & \xmark & \cmark \\
        \citet{abbaspourazad2023large} & 141 & 400 & \cmark & \cmark & \xmark  & \xmark  & \xmark  & \xmark  & \xmark  \\
        \citet{yuan2024self} & 100 & 15,700 & \xmark & \xmark & \cmark & \xmark & \xmark  & \xmark & \xmark \\
        \midrule
        \grayrow
        \textbf{\lsm(Ours)} & \textbf{165} & \textbf{40,000} & \xmark & \cmark & \cmark & \cmark & \cmark & \cmark & \cmark \\ 
    \bottomrule[1.5pt]
    \end{tabular}
    \label{tab:related_work}
    \vspace{8pt}
    \newline
    \footnotesize
    \ECGCircle: Electrocardiography, \PPGCircle: Photoplethysmography, \AccelerometerCircle: Accelerometer,\newline \EDACircle: Skin Conductance Level, \TempCircle: Skin Temperature, \AltCircle: Altimeter 
\vspace{-3pt}
\end{table}

\textbf{Time-Series Foundation Models.}
Wearable sensor data typically takes the form of multivariate time series. Foundation models for time-series signals have been trained and evaluated on data from domains such as energy use, transportation, finance, and climate.
TimeGPT \citep{garza2023timegpt} and Lag-Llama \citep{rasul2023lag} represented early versions of pretrained models for predicting time-series signals. Families of models for general-purpose time series analysis emphasize common properties present in many signals, even those from different sources \citep{goswami2024moment}. 
Recent efforts explore different model architectures \citep{das2023decoder} and scaling multiple data sources \citep{ansari2024chronos}, examing how language models can perform zero-shot reasoning~\citep{liu2023large, merrill2024language}.
Yet, time series from different domains can exhibit considerably different properties. Drawing inspiration from prior work, we focus on the analysis of \textit{sensory} time-series data, exploring scaling behavior, and interrogating whether they are consistent with other domains or show unique properties.

\textbf{Scaling Laws in Deep Learning.}
The scaling of computational resources, data volume, and model size has driven remarkable advancements in deep learning \citep{zhai2022scaling,kaplan2020scaling,xie2023data}. Recent investigations indicate that testing loss follows a power law relationship with each of these three resources when the other two are held constant \citep{kaplan2020scaling}. Power law behavior has been observed across various domains, including large language models \citep{kaplan2020scaling}, large vision models \citep{zhai2022scaling}, transfer learning \citep{hestness2017deep}, and multimodal models \citep{aghajanyan2023scaling}. In this work, we take a step further and investigate the scaling behavior of training foundation models for multimodal \textit{wearable sensor} data.

\section{Data for Wearable Foundation Models}
\label{sec:data_construction}

\subsection{Sensor Data and Processing}

Fitbit Sense 2 and Pixel Watch 2 have four \emph{sensors} of highest relevance to this work:  PPG, accelerometer, skin conductance, and altimeter/pressure sensors. From these input signals we compute a set of 26 \emph{signals} (features), as described in Table~\ref{tab:features} of Appendix~\ref{sec:dataset_additional_details}. Raw sensor data is not stored at this scale as it would impact the battery life and memory on the device. Thus, we focus on one-minute resolution signals. 

\EDACircle \textbf{Skin Conductance.} The EDA sensor is used to infer sympathetic arousal via changes in micro-sweat levels, a physiological response to stress. Two electrodes on the back of the device measure changes in skin conductance level (SCL), which varies with skin moisture levels. SCL data is sampled at 200 Hz, downsampled to 25 Hz via a boxcar filter, and smoothed with a 5-minute median and low-pass filters~\citep{mcduff2024does}. Per-minute tonic SCL slope and magnitude are then calculated. Due to the nature of the sensing mode operation, SCL data is only collected during non-exercise wake-periods.

\TempCircle \textbf{Skin Temperature.} A temperature sensor located near the wrist-facing surface of the device takes measurement every 10 seconds. Per-minute slope and magnitude values are calculated via linear regression. Skin temperature signals are available whenever EDA signals are available. 

\PPGCircle \textbf{Photoplethysmography.} A validated algorithm~\citep{nissen2022heart} is used to extract heart rate (HR) once per second from PPG. The per-minute HR data was calculated by taking the mean of the interpolated, per-second data across non-overlapping one-minute windows. An on-device peak detection algorithm identified PPG-based R-wave peaks from which RR intervals were calculated. RR intervals are susceptible to noise from multiple sources, including movement, electronic noise, and missed heartbeats. To account for noise, outliers were removed from each sliding 5-minute window using the median-filter based approach~\citep{natarajan2020heart}. The percentage of each 5-minute window with valid RR intervals are calculated and referred to as ``heart rate variability (HRV) percent good''.
Nine standard HRV metrics~\citep{shaffer2017overview} are calculated every minute over a sliding 5-minute window: RR mean, RR median, RR 20$^{th}$ percentile, RR 80$^{th}$ percentile, RR Shannon Entropy, RR differences Shannon Entropy, standard deviation of RR, root mean squared difference of RR intervals, and percentage of RR intervals greater than 30ms (PNN30).

\AccelerometerCircle \textbf{Accelerometer.} Ten signals are extracted from the 3-axis accelerometer: Jerk, steps, accelerometer log energy and energy ratio, covariance, number of zero crossings and standard deviation.
These signals are extracted by converting the 3-axis accelerometer to root mean squared magnitude (1D), and applying a high-pass filter (HPF) to the remove DC component.  In parallel, the 3-axis accelerometer signal is put through a second-order band-pass filter (BPF) and the principal component of the filtered 3-axis signal covariance matrix is calculated. In brief, jerk is a measure based on the time-derivative of the acceleration calculated from the principal component. It is the logarithm of the ratio of the absolute of the t=1 autocorrelation lag over the t=0 autocorrelation lag. Steps is a per-minute count of steps taken.  Log energy is the logarithm of the sum of the squared HPF signal over the window. Log energy ratio is the logarithm of the ratio of energy computed from principal-component over the magnitude of the HPF signal. Zero-crossing count is the number of crossings in the principal component. Kurtosis is the kurtosis of the BFP signal. 

\AltCircle \textbf{Altimeter.} The standard deviation of the altimeter (pressure sensor) measurements.

\begin{table}%
\caption{\textbf{Details of the datasets.} Summary of the demographic composition of our pretraining set and class distribution of our downstream set samples.}%
\vspace{-3pt}
\centering
\small
{
\renewcommand{\arraystretch}{1.15}
\subfigure[\textbf{Demographics of the pretraining set.}]{%
\label{tab:demographics}
\begin{tabular}{rcc}
\toprule
\textbf{Category}
& {\makecell[b]{\textbf{\# People}}}
& {\makecell[b]{\textbf{\%}}} \\ 
\midrule
\textbf{Sex} \quad \quad \quad \quad \quad~ Female & 110,780 & 67.0\%  \\ 
\quad Male & 53,895 & 32.6\% \\
\quad Not Specified & 415 & 0.3\%  \\ \hdashline
\textbf{Age} \quad \quad \quad \quad \quad \quad 18-39     & 55,653 & 33.7\% \\
\quad 40-59     & 75,627 & 45.8\% \\  
\quad 60-79     & 32,251 & 19.5\% \\  
\quad $\ge$80       & 1,548 & 0.9\% \\ \hdashline
\textbf{BMI} \quad \quad Healthy ($<$25)   & 57,015 & 34.5\% \\
\quad Overweight (25-30)   & 52,950 & 32.0\% \\
\quad Obese ($\ge$30)   & 54,727 & 33.1\% \\
\quad Not Specified  & 575 & 0.3\% \\
\hdashline
\quad \textbf{Total} & 165,090 & 100\% \\
  \bottomrule %
\end{tabular}
}
\hfill
\subfigure[\textbf{Class sample distribution of the downstream set.}]{%
\label{tab:activity_events}
\begin{tabular}{rcc}
    \toprule
        \textbf{Class}   & \textbf{\# Training} & \textbf{\# Testing} \\
        \midrule
        Exercise & 3,272 & 671\\
        Non-Exercise & 6,195 & 1,329 \\
        \hdashline
        \textbf{Total} & 9,467 & 2,000 \\ 
        \hline
        Biking & 1,191 & 412 \\ 
        Elliptical & 152 & 49 \\ 
        High Intensity Training & 332 & 104 \\ 
        Strength Training & 229 & 425 \\  
        Swimming & 2,332 & 441 \\ 
        Running & 1,860 & 315 \\ 
        Walking & 6,887 & 1,301 \\ 
        Weightlifting & 669 & 98 \\ 
        \hdashline
        \textbf{Total} & 14,372 & 3,262\\ 
        \bottomrule
        \end{tabular}}}
\vspace{-5pt}
\end{table}

All sensor signals were globally normalized (z-score) to remove differences in magnitude due to different units of measurement. As the masked autoencoder cannot process missing data, we imputed minutes that had missing values. Within each 300-minute window, missing data between valid data points was linearly interpolated, and leading missing minutes were backfilled.

\subsection{Building A Large Scale Pretraining Sensor Dataset}

To build the large dataset for our experiments we sampled wearable data from 165,090 subjects during the period January 1$^{st}$ 2023 to July 2$^{nd}$ 2024. The subjects wore Fitbit Sense 2 or Google Pixel Watch 2 devices and consented for their data to be used for research and development of new health and wellness products and services. We sub-selected from people wearing one of these devices as older device generations included fewer sensors. The subjects were asked for self-reported sex, age and weight. Table~\ref{tab:demographics} summarizes the characteristics of the pretraining data. All data were de-identified and not linked with any other information. To create a dataset that maximized the number of subjects we randomly sampled 10 5-hour windows of data from each subject, for a total of 8 million hours (6.6 million pretrain hours). We further explore the extremes of data scaling by experimenting with a subject-imbalanced 40 million hour pretraining dataset (see Appendix~\ref{sec:appendix_gen_tasks_scaling}).

The dataset was split 80-20 based on subjects into train-test splits. 
We then created several ``slices'' of the training set to conduct the scaling experiment.  The test set remains identical throughout all experiments. In the ``sample-scaling'' experiments we shuffled the training data and took N samples per experiment.  In the ``subject-scaling'' experiments we grouped the training data by subject identifier and took all samples from N subjects per experiment.

\section{Sensor Modeling Tasks}

\subsection{Generative Tasks} 

\begin{figure*}
    \centering
    \includegraphics[width=1\textwidth]{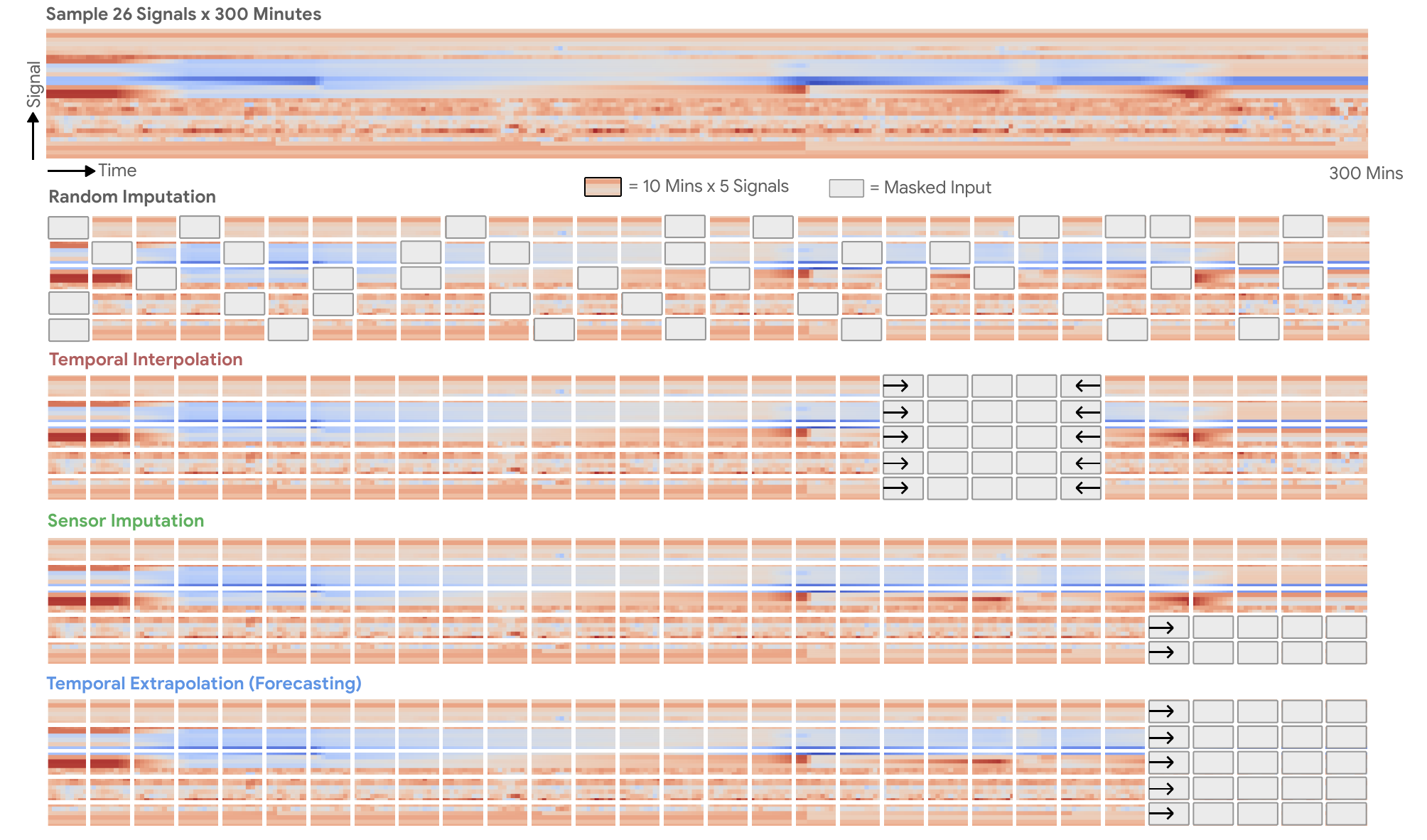}
    \caption{\textbf{Generative \lsm tasks and pretraining.} We define four distinct generative tasks: random imputation, temporal interpolation, signal/sensor imputation, and temporal extrapolation (forecasting). Random imputation was empirically chosen as the pretraining task.}
    \label{fig:tasks} 
\end{figure*}

We posit that defining generative tasks in the training of wearable sensor models may not only result in learned representations that are useful for downstream classification tasks, but also produce models that can impute missing or incomplete data (interpolate) and extrapolate future sensor values (forecast). To train the model and to test these capabilities we define several tasks (see Fig.~\ref{fig:tasks}).

\textbf{Random Imputation.} Our primary pretext task involves removing patches randomly from the input sample across the time-axis and signal-axis. During training this requires the model to infer missing values and make predictions based on partial input.

\textbf{Temporal Interpolation.} Sensor inputs can be missing for a number of reasons. Devices need to be removed from the wrist for charging, and certain sensors might be turned off for periods to save on battery life~\citep{mcduff2024does}. Interpolation of sensor data is an important and necessary step for many algorithms (see Fig.~\ref{fig:tasks}). In this task we test the model's ability to fill gaps in the data where all sensor data is missing for a period of time, usually between two observations. 

\textbf{Sensor Imputation.} Sensor imputation refers to the process of inferring a subset of partially missing sensor-streams, from other continuously online sensing modalities. By leveraging correlations between different physiological signals, sensor imputation ensures that insights can be derived even when some sensor modalities are absent, enhancing the overall versatility and capabilities of multi-sensor systems. Under the constraints of hardware limitations (battery, wireless connectivity, etc.), sensor imputation can enable the delivery of more realistic metrics to the user (e.g., step count, average resting heart rate) even if when sensors are not continuously online.

\textbf{Temporal Extrapolation (Forecasting).} A more challenging task than interpolation is extrapolation of sensor values forward in time. Temporal extrapolation involves predicting future sensor measurements. The ability to anticipate future physiological states based on current and historical data has applications in areas such as health interventions, where extrapolation can be used to schedule recovery times, detect early signs of fatigue, predict wake-up times, and detect anomalies. Accurate signal extrapolation is a key task that can empower wearable devices to provide more just-in-time, proactive, and personalized health recommendations.

\subsection{Discriminative Tasks}
\label{sec: discriminative_task}

Discriminative tasks focus on classifying or identifying specific activities, states, or conditions based on sensor data. These tasks are essential for translating raw sensor inputs into actionable, personalized, and relevant feedback. Two exemplary tasks are considered here.

\textbf{Exercise Detection.} Exercise detection identifies when a user is exercising, enabling real-time feedback and performance tracking. This task involves recognizing exercise events from continuous sensor data, allowing devices to log workout sessions, track progress, and provide personalized recommendations. Additionally, detecting exercise unlocks related experiences, such as identifying exercise types, marking session start times, or tracking post-exercise feedback. We developed a dataset with windows of user-labeled exercise and non-exercise events (see Table~\ref{tab:activity_events}).

\textbf{Activity Recognition.} Activity recognition is the process of classifying different user activities such as biking, running, or walking, based on the patterns detected in sensor data. This allows wearable devices to monitor daily routines accurately, providing insights into fitness levels, activity trends, and overall health. Effective activity recognition enables applications like fitness tracking, lifestyle monitoring, and personalized coaching. Our dataset includes eight user-labeled activities: \emph{Biking}, \emph{Elliptical}, \emph{High-Intensity Interval Training (HIIT)}, \emph{Strength Training}, \emph{Swimming}, \emph{Running}, \emph{Walking}, and \emph{Weightlifting}. 

\section{Experiments \& Results}
\label{sec:experiment_results}
\subsection{Training Procedures}

We pretrain wearable foundation models on a diverse collection of multimodal sensor data from 80\% of the 165,090 subjects as described in Table \ref{tab:demographics}. Each sample is processed as a two-dimensional matrix of 26 signals by 300 minutes (see Fig. \ref{fig:tasks}). Our primary pretraining objective is to optimize the masked signal reconstruction loss (i.e., mean squared error), averaged over randomly masked patches from the input sequences \citep{he2022masked}. The primary performance metric is the mean squared error on the held-out test set, evaluated across all the normalized signals.

We train our models on Google v5e TPUs with a total batch size of 4096 across 50,000 training steps. The training process uses the AdamW optimizer with a base learning rate of $5e-3$ and weight decay set to $1e-4$. A linear warm-up schedule is applied for the first 2,500 steps, followed by a cosine learning rate decay to zero. All pretraining experiments use an 0.8 masking ratio (masking out random patches that cover 80\% of the total input signals). Additional details on implementation and hyperparameters can be found in Appendix \ref{sec:appendix_hyperparas}.

\begin{figure*}[t!]
    \centering
    \includegraphics[width=\textwidth]{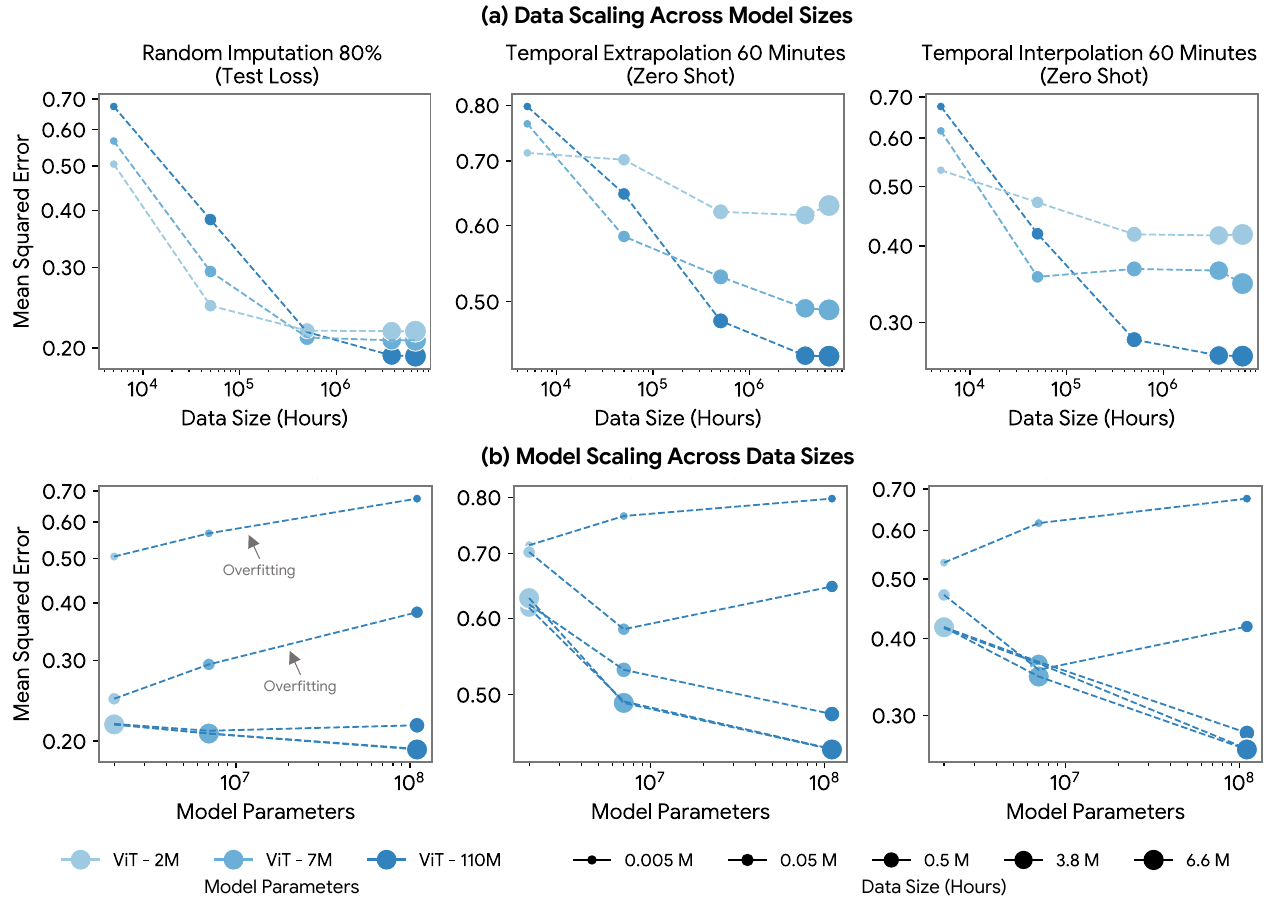}
    \caption{\textbf{Scaling performance of \lsm.} We show performance on \textit{generative tasks} across varying data and model sizes. \lsm begins to saturate at approximately $10^7$ hours of data. The effects of scaling are more pronounced in imputation, interpolation, and extrapolation tasks. Results indicate that as model size increases, significantly larger data volumes are required to prevent overfitting.}
    \label{fig:gen_task_data_model_scaling} 
    \vspace{-4pt}
\end{figure*}

\subsection{Results \& Discussion}

\textbf{Do scaling laws apply to wearable data?} We present the Pareto front of the reconstruction loss and downstream performance as a function of \textit{compute scaling} (see Fig. \ref{fig:teaser}). The front highlights the models with optimal compute allocation across model size, data size and training duration. Over multiple orders of magnitude of compute, the relationship between compute and performance follows a power-law ($L = aC^{b}$), resulting in a nearly linear trend on the log-log plot. However, we observe a saturation effect at the upper end of the compute spectrum, where the largest models do not asymptotically approach zero error. This behavior has also been observed for scaling language models \citep{henighan2020scaling} and vision transformers \citep{zhai2022scaling}; therefore, we add an additive constant $c$ to model this saturation effect: $L = aC^{b} + c$.

\begin{figure}[!t]
\subfigure[Scaling Subjects]{\includegraphics[width=0.31\linewidth]{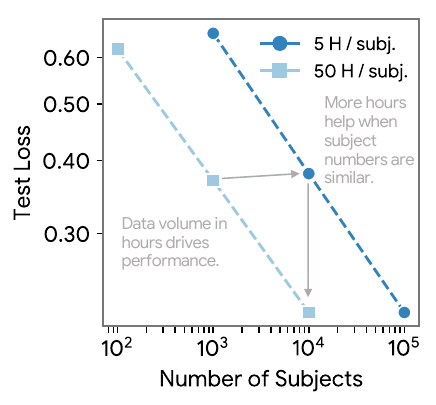} \label{fig:data_subjects_scaling}}
\subfigure[Scaling on Downstream Tasks]{\includegraphics[width=0.33\linewidth]{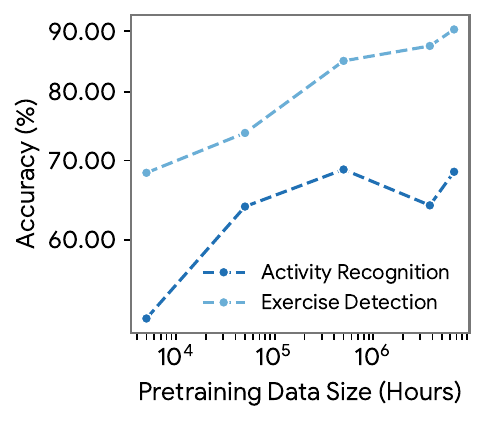} \label{fig:data_scaling_dis_task}}
\subfigure[Learning Efficiency]{\includegraphics[width=0.32\linewidth]{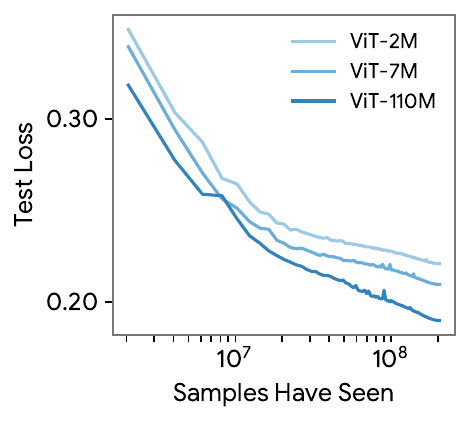} \label{fig:model_efficiency}}
 \caption{\textbf{Analysis on scaling \lsm.} \textbf{(a)} Total number of hours is more important than total number of subjects. \textbf{(b)} Data scaling on discriminative tasks with ViT-110M. \textbf{(c)} Larger models are more sample efficient.}
\vspace{-6mm}
\end{figure}

We illustrate \textit{data scaling} across various model sizes (Fig.~\ref{fig:gen_task_data_model_scaling}\color{gray}(a)\color{black}). Performance improves monotonically to approximately $10^5$ data hours, beyond which the rate of improvement diminishes, particularly around $10^7$ hours. We validated that scaling beyond $10^7$ hours yields minimal benefits by training with 40 million hours (see Appendix~\ref{sec:appendix_gen_tasks_scaling}). Consequently, results in Table~\ref{tab:task_results} are pretrained with 6.6 million hours of data. Larger models, especially the ViT-110M, continue to benefit from data scaling, showing substantial gains when training on over 1 million hours of data. These observations underscore the large data requirements needed to fully exploit the capacity of larger models, which are far greater than those required by smaller models. A similar trend is observed in discriminative tasks (Fig.~\ref{fig:data_scaling_dis_task}). We further note that these trends are based on minutely aggregated wearable data; raw sensor signals are traditionally collected at substantially higher sampling frequencies and it is possible that feature extraction on more fine-grained sensor data may require even larger models.

\textit{Model scaling} results as a function of data size demonstrate that as both model size and dataset size are scaled, sufficient data is essential to prevent overfitting (Fig.~\ref{fig:gen_task_data_model_scaling}\color{gray}(b)\color{black}). Models trained on smaller datasets exhibit limited generalization capacity, whereas scaling up to $10^8$ parameters results in significant gains in test loss and generative zero-shot performance. These findings highlight the need to align model size with adequate data to fully leverage the model's representational power. Our experiments also show larger models are more sample efficient as illustrated in Fig.~\ref{fig:model_efficiency}. 

By scaling compute, data, and model size together, \lsm achieves improvements of 16\% to 23\% in temporal interpolation MAE and 20\% to 21\% in extrapolation MAE across five time durations as compared to the best baseline method (Table \ref{tab:generative_task_results}). Additionally, \lsm outperforms baselines in exercise detection and 8-class activity recognition over the supervised baseline by 27\% / 29\% in accuracy and 57\% / 54\% in mAP, as detailed in Table \ref{tab:discriminative_task_results_activity}. Our baseline approaches are commonly used in existing sensor algorithms~\citep{gershon2016daily,van2023missing}. More scaling results can be found in Appendix \ref{sec:additional_results}.

\textbf{Is scaling subjects or wearable data hours per subject more helpful?} As shown in Fig.~\ref{fig:data_subjects_scaling}, when training using the same total number of wearable signal hours, reducing the number of subjects (but drawing more hours per subject) can yield similar performance. This suggests that total number of hours rather than number of subjects drives gains. One possible hypothesis to explain this effect is that the diversity of activities per subject (as reflected by the increase in hours per subject) plays a crucial role. Alternatively, subject diversity may become more important when learning representations if we scale up the data sample size from 5 hours (e.g., 7 days vs. 5 hours). As each subject has a finite number of hours, to maximize model generalization, it is important to scale both the number of subjects and the wearable data hours per subject simultaneously. While temporal data is crucial for capturing intra-subject variability, increasing the number of subjects introduces valuable inter-subject diversity. Therefore, scaling both dimensions—subjects and hours—together is essential to fully leverage the model's capacity and improve performance across tasks.

\textbf{Can wearable foundation models impute the past and predict the future?} As shown in Fig.~\ref{fig:gen_task_data_model_scaling}, scaling laws apply to all imputation, interpolation, and extrapolation tasks, with larger models and more data resulting in improved performance. The utility of \lsm is further emphasized in Table~\ref{tab:generative_task_results}. However, despite these quantitative gains, the qualitative results in Fig.~\ref{fig:reconstruction_examples} of Appendix~\ref{appendix-sec:reconstructions} reveal that these tasks remain highly challenging. Imputing large portions of missing data, especially over extended time intervals, often leads to degraded accuracy, with performance deteriorating as the missing data window increases. Similarly, extrapolation further into the future (e.g., several hours ahead) introduces significant uncertainty, making it difficult to predict fine-grained physiological or behavioral patterns. These findings suggest that while scaling helps improve generative capabilities, substantial challenges remain, particularly in handling long-range dependencies and large data gaps.

\textbf{Are wearable foundation models label efficient on discriminative tasks?} Our experiments on probing, fine-tuning, and few-shot learning for activity indicate that wearable foundation models are highly label efficient. As shown in Table~\ref{tab:discriminative_task_results_exercise} and \ref{tab:discriminative_task_results_activity}, the performance of the fine-tuned \lsm consistently outperforms supervised baselines. A confusion matrix of the best performing model is shown in Fig.~\ref{fig:confusion_matrix}. As shown in Table~\ref{tab:few_shot_data} of Appendix~\ref{sec:appendix_dis_tasks_scaling}, even in the low-data regime (e.g., 5-shot, 10-shot), foundation models demonstrate strong generalization capabilities, achieving significantly lower error rates compared to models trained from scratch or with limited supervision. As the number of labeled examples increases, the performance gap widens, with foundation models leveraging pretraining to more effectively transfer learned representations to downstream tasks. T-distributed Stochastic Neighbor Embeddings (t-SNE) plots show the impact of pretraining on more data and fine-tuning are shown in Appendix~\ref{appendix:feature_embeddings} (Fig.~\ref{fig:activity_tsne}).

\begin{table}%
\caption{\textbf{Comparisons of \lsm and competing methods on generative and discriminative tasks.}}
\vspace{-0.2cm}
\label{tab:task_results}
\centering
\small
\subfigure[\textbf{Generative Task Results}]{%
    \label{tab:generative_task_results}
    \resizebox{\textwidth}{!}{%
    \begin{tabular}{lccccc}
    \toprule
        \textbf{Task + Method} & \multicolumn{3}{c}{\textbf{Error (MAE / MSE)}} \\
        \midrule
        \textbf{Temporal Interpolation}  & \textbf{10 mins} & \textbf{20 mins} & \textbf{30 mins} & \textbf{60 mins} & \textbf{120 mins} \\
        \midrule
         \textsc{Mean} & 0.36 / 0.42 & 0.36 / 0.43 & 0.37 / 0.44 & 0.38 / 0.46 & 0.39 / 0.49 \\
         \textsc{Nearest Neighbor} & 0.21 / 0.29 & 0.26 / 0.37 & 0.28 / 0.42 & 0.33 / 0.51 & 0.38 / 0.62 \\
         \textsc{Linear Interp.} & 0.19 / 0.23 & 0.23 / 0.30 & 0.26 / 0.34 & 0.30 / 0.42 & 0.36 / 0.51 \\
         \grayrow
         \textsc{\lsm(MAE)} & \textbf{0.16} / \textbf{0.14} & \textbf{0.19} / \textbf{0.18} & \textbf{0.20} / \textbf{0.21} & \textbf{0.24} / \textbf{0.26} & \textbf{0.29} / \textbf{0.33} \\
         \midrule
        \textsc{Gains over Interp.} &
        \textcolor{darkgreen}{\texttt{+}\textbf{16\% / 39\%}} &
        \textcolor{darkgreen}{\texttt{+}\textbf{17\% / 40\%}} &
        \textcolor{darkgreen}{\texttt{+}\textbf{23\% / 38\%}} & \textcolor{darkgreen}{\texttt{+}\textbf{20\% / 38\%}} & \textcolor{darkgreen}{\texttt{+}\textbf{19\% / 33\%}} \\
        \midrule
        \textbf{Temporal Extrapolation} & \textbf{10 mins} & \textbf{20 mins} & \textbf{30 mins} & \textbf{60 mins} & \textbf{120 mins} \\
        \midrule
         \textsc{Mean} & 0.48 / 0.66 & 0.48 / 0.65 & 0.47 / 0.65 & 0.47 / 0.64 & 0.45 / 0.64 \\
         \textsc{Nearest Neighbor} & 0.35 / 0.52 & 0.40 / 0.62 & 0.43 / 0.68 & 0.47 / 0.76 & 0.48 / 0.81 \\
         \textsc{Linear Interp.} & 0.35 / 0.52 & 0.40 / 0.62 & 0.43 / 0.68 & 0.47 / 0.76 & 0.48 / 0.81 \\
         \grayrow
         \textsc{\lsm(MAE)} & \textbf{0.28} / \textbf{0.31} & \textbf{0.32} / \textbf{0.37} & \textbf{0.34} / \textbf{0.40} & \textbf{0.37} / \textbf{0.44} & \textbf{0.38} / \textbf{0.47}\\
         \midrule
         \textsc{Gains over Interp.} &
         \textcolor{darkgreen}{\texttt{+}\textbf{20\% / 40\%}} &
         \textcolor{darkgreen}{\texttt{+}\textbf{20\% / 40\%}} &
         \textcolor{darkgreen}{\texttt{+}\textbf{21\% / 23\%}} & \textcolor{darkgreen}{\texttt{+}\textbf{21\% / 31\%}} & \textcolor{darkgreen}{\texttt{+}\textbf{21\% / 27\%}} \\
         
        \midrule
        \textbf{Sensor Imputation} & \textbf{10 mins} & \textbf{20 mins} & \textbf{30 mins} & \textbf{60 mins} & \textbf{120 mins} \\
        \midrule
        \textsc{Mean} & 0.36 / 0.42 & 0.36 / 0.43 & 0.37 / 0.43 & 0.38 / 0.45 & 0.39 / 0.49 \\
        \textsc{Nearest Neighbor} & 0.21 / 0.29 & 0.26 / 0.37 & 0.28 / 0.42 & 0.33 / 0.51 & 0.38 / 0.62 \\
        \textsc{Linear Interp.} & 0.19 / 0.23 & 0.23 / 0.30 & 0.26 / 0.34 & 0.30 / 0.42 & 0.36 / 0.51\\
        \grayrow
        \textsc{\lsm(MAE)} & \textbf{0.15} / \textbf{0.11} & \textbf{0.15} / \textbf{0.12} & \textbf{0.16} / \textbf{0.13} & \textbf{0.17} / \textbf{0.15} & \textbf{0.19} / \textbf{0.17} \\
        \midrule
        \textsc{Gains over Interp.} & \textcolor{darkgreen}{\texttt{+}\textbf{21\% / 52\%}} & 
        \textcolor{darkgreen}{\texttt{+}\textbf{35\% / 60\%}} &
        \textcolor{darkgreen}{\texttt{+}\textbf{38\% / 62\%}} &
        \textcolor{darkgreen}{\texttt{+}\textbf{43\% / 64\%}} &
        \textcolor{darkgreen}{\texttt{+}\textbf{47\% / 67\%}} \\
        
        \midrule
         
         \bottomrule[1.2pt]
        \end{tabular}
        }
}
\subfigure[\textbf{Exercise Detection.}] {%
    \label{tab:discriminative_task_results_exercise}
    \resizebox{0.47\textwidth}{!}{%
    \begin{tabular}{llcc}
    \toprule       
        \textbf{Pretrain} & \textbf{Probe/FT} & \textbf{Acc.} & \textbf{mAP} \\
        \midrule
         - & \textsc{Supervised} & 70.9 & 61.7 \\
         \hdashline
         \textsc{MSN} & Linear Probe & 67.6 & 60.0 \\
         \textsc{DINO}  & Linear Probe & 66.0 & 57.0 \\
         \textsc{SimCLR}  & Linear Probe & 66.5 & 51.5 \\
         \textsc{\lsm(MAE)} & Linear Probe & 84.7 & 89.0 \\
         \hdashline
         \textsc{MSN} & Fine-tune & 76.7 & 74.6 \\
          \textsc{DINO}  & Fine-tune & 78.2 & 80.3  \\
         \textsc{SimCLR}  & Fine-tune & 74.9 & 66.6 \\
         \grayrow
         \textsc{\lsm(MAE)} & Fine-tune & \textbf{90.3} & \textbf{97.0} \\
         \midrule
         \multicolumn{2}{l}{\textsc{Gain over Supervised}} & \textcolor{darkgreen}{\texttt{+}\textbf{27\%}} & \textcolor{darkgreen}{\texttt{+}\textbf{57\%}} \\
        \bottomrule[1.2pt]
        \end{tabular}
        }}
\hfill
\subfigure[\textbf{Activity Recognition.}]{%
    \label{tab:discriminative_task_results_activity}
    \resizebox{0.47\textwidth}{!}{%
    \begin{tabular}{llcc}
    \toprule       
        \textbf{Pretrain} & \textbf{Probe/FT} & \textbf{Acc.} & \textbf{mAP} \\
        \midrule
         - & \textsc{Supervised} & 53.2 & 33.4 \\
         \hdashline
         \textsc{MSN} & Linear Probe & 44.6 & 24.0 \\
         \textsc{DINO} & Linear Probe & 50.3 & 26.0 \\
         \textsc{SimCLR} & Linear Probe & 45.3 & 20.8  \\
         \textsc{\lsm(MAE)} & Linear Probe & 49.4 & 24.6 \\

         \hdashline
         \textsc{MSN} & Fine-tune & 62.5 & 43.4 \\
         \textsc{DINO} & Fine-tune & 66.2 & 46.3 \\
         \textsc{SimCLR} & Fine-tune & 67.3 & 46.0 \\
         \grayrow
         \textsc{\lsm(MAE)} & Fine-tune & \textbf{68.5} & \textbf{51.4}  \\
         \midrule
         \multicolumn{2}{l}{\textsc{Gain over Supervised}} &  \textcolor{darkgreen}{\texttt{+}\textbf{29\%}} & \textcolor{darkgreen}{\texttt{+}\textbf{54\%}} \\

        \bottomrule[1.2pt]
        \end{tabular}
        }}
        \vspace{-0.1cm}
        \begin{flushleft}
        \fontsize{8}{8}\selectfont 
        All neural methods, including the supervised method, utilize a ViT-Base (110M) backbone. Relevant methods are pretrained with 6.6 million hours of data. In the sensor imputation task, we randomly mask 67\% of the sensor modalities. MSN~\citep{assran2022masked}, DINO~\citep{caron2021emerging}, SimCLR~\citep{chen2020simple}, MAE~\citep{he2022masked}.
        \end{flushleft}
        \vspace{-0.7cm}
\end{table}

\subsection{Further Analysis and Ablation Studies}

\textbf{Ablation of Model Design Choice (Appendix \ref{sec:appendix-hyperparameter-experiments})}. We analyze the impact of design choices on \lsm performance, including masking ratios and strategies, signal orders, patch sizes, and model sizes.

\textbf{Qualitative Analysis of \lsm (Appendix \ref{appendix:feature_embeddings} \& \ref{appendix-sec:reconstructions})}. We further explore the learned feature embeddings to assess their sensitivity to personally identifiable features (e.g., age and gender) and examine the reconstruction quality of the signals.

\section{Limitations \& Future Work}
\label{sec:limitation_and_future_work}
Our experiments indicate promising opportunities in scaling wearable sensor models but also highlight several unresolved questions. Notably, we observe saturation in scaling laws with a dataset size of $10^7$ hours and model sizes in 100 millions. We attribute this to three factors: (1) the current pretraining task may not be sufficiently scalable, and decoder-only approaches might better leverage data rather than filling masked inputs; (2) the dataset construction lacks sufficient challenge, and extending the sensor context window from 5 hours to a day or even a week could introduce more complexity that enables the model to learn longer time dependency relationships; (3) our data cleaning process was minimal, and increasing data diversity, akin to large-scale language model training, could significantly enhance model generalization. For example, while our dataset spanned all four seasons, there was an imbalance in temporal coverage, with two years of data from January to June but only a single year from July to December. This uneven distribution could bias the model towards activities more common in the earlier part of the year.

A key characteristic of wearable sensor data is its inherent missingness. Handling missing data in both pretraining and downstream tasks remains an open question. While we used imputation for this study, a more principled approach would involve designing models that naturally account for missing data without introducing imputation biases. The nature of missing data in wearable sensors often correlates with real-world events (e.g., charging the device, loose fitting), which can mean that data is missing not at random (MNAR). Understanding these factors and designing methods to handle them robustly remains an important direction for future work. Lastly, we acknowledge the lack of comprehensive evaluation on more discriminative tasks. Future work will expand the dataset to include a broader range of classification and regression tasks, which will provide a more thorough demonstration of the benefits of our pretrained models.

\section{Broader Impact}

Wearable sensors have been shown to have a positive effect on health and well-being, promoting physical activity, sleep and have potential to surface unseen or unperceived actionable health information. Foundation models increase the potential value of these data for the above applications and hold promise for enabling new insights and opportunities to improve health. 

We support open science principles and the value of open data for scientific research; however, we have to balance these considerations with the privacy of the participants and protection of their health data. Although the training data could be de-identified, some of the data streams could not be fully anonymized.
We recognize that the inability to share data of this kind is a limitation; however we believe that the results enable us to share valuable insights to the community. 

Meanwhile, \lsm serves as the stepping stone towards generating large-scale, realistic synthetic datasets. These synthetic data could mimic real-world sensor patterns without compromising participant privacy and offer a promising resource for cross-institutional research collaboration.
By facilitating data sharing in this way, we can overcome the current limitations in data availability and unlock new opportunities for collaborative insights and advancements for the community.

\section{Conclusion}

We present \lsm, a large multimodal foundation model trained on 40 million hours of wearable sensor data from over 165,000 individuals, establishing scaling laws for sensor models. \lsm significantly improves performance across generative tasks such as imputation, interpolation, and extrapolation, as well as discriminative tasks like exercise detection and activity recognition. Our results demonstrate that scaling data, model size, and compute leads to substantial gains in generalization and efficiency. \lsm highlights the potential of scaling wearable sensor models for real-world health applications, enabling more robust and efficient downstream tasks.

\bibliography{main_arxiv}

\appendix

\section*{\LARGE Appendix}

\startcontents[appendices]

\section{Model Design Choices and Ablations}
\label{sec:appendix-hyperparameter-experiments}

We perform ablations on the configurations used for our masked autoencoder \lsm design. Following the convention of previous works \citep{he2022masked, huang2022masked}, we explore masking ratio, masking strategies, patch sizes, and model sizes. Uniquely, we explore the ordering of sensor signals, as these signals do not share the same explicit ordered dependencies as exist in images and audio spectrograms. For all experiments we employ random masking, a 0.8 masking ratio, ordered sensor signal order, a patch size of 10x5, and a \lsm-Base (110M) backbone, unless otherwise specified.

\subsection{Selecting a Masking Ratio} 

Selecting the appropriate masking ratio is critical for ensuring effective representation learning in our sensor MAE training. We explore different masking ratios, ranging from 30\% to 90\%, to evaluate their impact on reconstruction quality and model generalization. We find that a masking ratio of 80\% yields the best performance on temporal interpolation and extrapolation as shown in Table \ref{tab:masking_ratio_ablation}).

\begin{table}[h!]
\setlength{\tabcolsep}{2.5pt}
\vspace{-1.5pt}
\small
\begin{center}
\begin{minipage}[t]{\textwidth}
\centering
\caption{\small \textbf{Ablation study of masking ratios.}}
\label{tab:masking_ratio_ablation}

\adjustbox{max width=\textwidth}{
\begin{tabular}{lcccc}
\toprule[1.5pt]
\multirow{2}{*}{\textbf{Mask Ratio}} & \multicolumn{2}{c}{\textbf{Interpolation 60 mins}} & \multicolumn{2}{c}{\textbf{Extrapolation 60 mins}} \\
\cmidrule(r){2-3}\cmidrule(r){4-5}
& \textbf{MAE} & \textbf{MSE} & \textbf{MAE} & \textbf{MSE} \\  
\midrule\midrule
\textsc{0.3} & 0.29 & 0.31 & 0.38 & 0.47 \\[1.2pt]
\textsc{0.4} & 0.35 & 0.39 & 0.38 & 0.45 \\[1.2pt]
\textsc{0.5} & 0.25 & 0.27 & 0.39 & 0.46 \\[1.2pt]
\textsc{0.6} & 0.44 & 0.57 & 0.38 & 0.45 \\[1.2pt]
\textsc{0.7} & 0.40 & 0.51 & \textbf{0.37} & \textbf{0.44} \\[1.2pt]
\grayrow
\textsc{0.8} & \textbf{0.24} & \textbf{0.26} & \textbf{0.37} & \textbf{0.44} \\[1.2pt]
\textsc{0.9} & 0.31 & 0.33 & 0.40 & 0.49 \\[1.2pt]
\bottomrule[1.5pt]
\end{tabular}}
\end{minipage}%
\end{center}
\end{table}

\subsection{Selecting a Masking Strategy}
\begin{figure*}[h!]
    \centering
    \includegraphics[width=\textwidth]{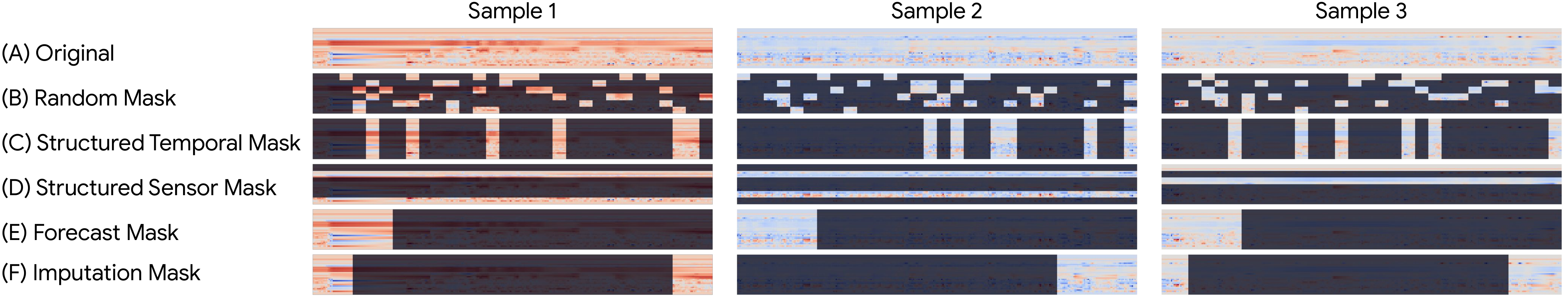}
    \caption{\textbf{\lsm MAE pretrain masking strategies.} All strategies employ a masking ratio of 0.8. \textbf{(A)}: original, unmasked sensor image, \textbf{(B)}: random masking, \textbf{(C)}: structured temporal masking, \textbf{(D)}: structured sensor masking, \textbf{(E)}: temporal extrapolation masking, \textbf{(F)}: temporal interpolation masking. Both random and structured temporal masking enable strong down-stream performance. We select random masking for all scaling experiments and evaluations.}
    \label{fig:mask_strategies} 
    \vspace{-4pt}
\end{figure*}

To train a wearable foundation model effective for both generative and discriminative tasks, mask-based pretraining proves superior to contrastive pretraining. Choosing the right masking strategy is crucial, as it directly influences the quality of the learned embeddings and the model's generalizability. In Table \ref{tab:masking_strategy_ablation}, we systematically compare five different masking strategies and demonstrate that random masking consistently yields the best performance across the two primary generative tasks. Example visualizations of these masking strategies can be seen in Fig. \ref{fig:mask_strategies}.

\begin{table}[h!]
\setlength{\tabcolsep}{2.5pt}
\vspace{-1.5pt}
\small
\begin{center}
\begin{minipage}[t]{\textwidth}
\centering
\caption{\small \textbf{Ablation study of masking strategies.}}
\label{tab:masking_strategy_ablation}

\adjustbox{max width=\textwidth}{
\begin{tabular}{lcccc}
\toprule[1.5pt]
\multirow{2}{*}{\textbf{Mask Strategy}} & \multicolumn{2}{c}{\textbf{Interpolation 60 mins}} & \multicolumn{2}{c}{\textbf{Extrapolation 60 mins}} \\
\cmidrule(r){2-3}\cmidrule(r){4-5}
& \textbf{MAE} & \textbf{MSE} & \textbf{MAE} & \textbf{MSE} \\  
\midrule\midrule

\grayrow
\textsc{Random} & \textbf{0.24} & \textbf{0.26} & 
\textbf{0.37} & \textbf{0.44} \\[1.2pt]
\textsc{Structured (Temporal)} & \textbf{0.24} & \textbf{0.26} &
\textbf{0.37} & \textbf{0.44} \\[1.2pt]
\textsc{Structured (Sensor)} & 0.54 & 0.71 & 0.53 & 0.73 \\[1.2pt]
\textsc{Temporal Interpolation} & 0.41 & 0.48 & 0.52 & 0.66 \\[1.2pt]
\textsc{Temporal Extrapolation} & 0.43 & 0.51 & 0.51 & 0.64 \\[1.2pt]
\bottomrule[1.5pt]
\end{tabular}}
\end{minipage}%
\end{center}
\end{table}

\subsection{Selecting a Sensor Signal Order} 
For multimodal sensor data, the order in which signals are processed by the model can impact performance. Specifically, for architectures, such as vision transformers, that take patched inputs, the clustering of signals in patches may have a profound impact on the learned representation. We evaluate ordering the sensor signals by: (a) sensor types (as in Table~\ref{tab:features}), (b) randomized order (repeated with several random seeds) and (c) interleaving signals with uncorrelated signals. Cross correlation matrices are shown in Fig.~\ref{fig:feature_correlations}. We find that ordering by clustering sensor type generally yields better results (see Table \ref{tab:sensor_order_ablation}), particularly when dealing with heterogeneous sensor modalities like accelerometry, electrodermal activity (EDA), and heart rate. This order allows the model to leverage specific sensor characteristics more effectively, improving performance on downstream tasks.

\begin{table}[h!]
\setlength{\tabcolsep}{2.5pt}
    \caption{\small \textbf{Ablation study of sensor orders.}}
\vspace{-1.5pt}
\label{tab:sensor_order_ablation}
\small
\begin{center}
\begin{minipage}[t]{\textwidth}
\centering
\adjustbox{max width=\textwidth}{
\begin{tabular}{lcccc}
\toprule[1.5pt]
\multirow{2}{*}{\textbf{Sensor Order}} & \multicolumn{2}{c}{\textbf{Interpolation 60 mins}} & \multicolumn{2}{c}{\textbf{Extrapolation 60 mins}} \\
\cmidrule(r){2-3}\cmidrule(r){4-5}
& \textbf{MAE} & \textbf{MSE} & \textbf{MAE} & \textbf{MSE} \\  
\midrule\midrule
\grayrow
\textsc{Clustered} & \textbf{0.24} & \textbf{0.26} & \textbf{0.37} & \textbf{0.44} \\[1.2pt]
\textsc{Randomized (N=5)} & 0.28 & 0.32 & 0.38 & 0.45 \\[1.2pt]
\textsc{Max Entropy} & 0.30 & 0.34 & 0.45 & 0.55 \\[1.2pt]
\bottomrule[1.5pt]
\end{tabular}}
\end{minipage}%
\end{center}
\end{table}

\subsection{Selecting a Patch Size.} Patch size in our pretraining is defined by time steps and the number of sensor features per patch, both impacting model capacity and computation (gFlops). In contrast to previous works ~\citep{he2022masked, huang2022masked} we expensively sweep across both dimensions of the input. This is critical for sensor models, where both dimensions, of time and features, share unique correlations and dependencies along their corresponding axis.

A time-step of 10 minutes strikes the best balance with low gFlops (15.94) and strong performance (MAE of 0.24 for imputation, 0.37 for forecasting) (See Table \ref{tab:patch_size_ablation}). Similarly, increasing features per patch shows that five features per patch achieves the best trade-off between accuracy and computational cost, outperforming both smaller (10x1) and larger patches (10x26). Thus, a moderate patch size of 10 minutes by 5 features is what we select.

As a patch-size of 10-minutes x 5-sensors (10x5) cannot evenly patch a input sensor-image of 300-minutes x 26-sensors (300x26), we zero-pad the sensor dimension to 30 resulting in an 300-minute x 30-feature (300x30) input sensor-image.

\begin{table}[h!]
\setlength{\tabcolsep}{2.5pt}
\caption{\small \textbf{Ablation study of patch sizes.}}
\vspace{-1.5pt}
\label{tab:patch_size_ablation}
\small
\begin{center}
\begin{minipage}[t]{0.49\textwidth}
\centering
\adjustbox{max width=\textwidth}{
\begin{tabular}{lccccc}
\multicolumn{6}{l}{(a) \textbf{Sweep across time-steps per patch.} (5 feats. per patch)} \\
\toprule[1.5pt]
\multirow{2}{*}{\textbf{Patch Size}} & \multirow{2}{*}{\textbf{gFlops}} & \multicolumn{2}{c}{\textbf{Interpolation 60 mins}} & \multicolumn{2}{c}{\textbf{Extrapolation 60 mins}} \\
\cmidrule(r){3-4}\cmidrule(r){5-6}
& & \textbf{MAE} & \textbf{MSE} & \textbf{MAE} & \textbf{MSE} \\  
\midrule\midrule
\textsc{5\textcolor{gray}{x5}} & 33.09 & 0.34 & 0.41 & \textbf{0.37} & 0.46 \\[1.2pt]
\grayrow
\textsc{10\textcolor{gray}{x5}} & 15.94 & \textbf{0.24} & \textbf{0.26} & \textbf{0.37} & \textbf{0.44} \\[1.2pt]
\textsc{20\textcolor{gray}{x5}} & 7.82 & 0.26 & 0.28 & 0.41 & 0.48 \\[1.2pt]
\textsc{30\textcolor{gray}{x5}} & \textbf{5.18} & 0.28 & 0.30 & \textbf{0.37} & \textbf{0.44} \\[1.2pt]
\bottomrule[1.5pt]
\end{tabular}}
\end{minipage}%
\hfill
\begin{minipage}[t]{0.49\textwidth}
\centering
\adjustbox{max width=\textwidth}{
\begin{tabular}{lccccc}
\multicolumn{6}{l}{(b) \textbf{Sweep across features per patch.} (10 mins. per patch)} \\
\toprule[1.5pt]
\multirow{2}{*}{\textbf{Patch Size}} & \multirow{2}{*}{\textbf{gFlops}} & \multicolumn{2}{c}{\textbf{Interpolation 60 mins}} & \multicolumn{2}{c}{\textbf{Extrapolation 60 mins}} \\
\cmidrule(r){3-4}\cmidrule(r){5-6}
& & \textbf{MAE} & \textbf{MSE} & \textbf{MAE} & \textbf{MSE} \\  
\midrule\midrule
\textsc{\textcolor{gray}{10x}1} & 77.83 & 0.30 & 0.33 & 0.43 & 0.53 \\[1.2pt]
\textsc{\textcolor{gray}{10x}2} & 36.07 & 0.30 & 0.33 & 0.45 & 0.53 \\[1.2pt]
\grayrow
\textsc{\textcolor{gray}{10x}5} & 15.94 &  \textbf{0.24} &  \textbf{0.26} &  \textbf{0.37} &  \textbf{0.44} \\[1.2pt]
\textsc{\textcolor{gray}{10x}10} & 7.82 & 0.33 & 0.38 & 0.45 & 0.55 \\[1.2pt]
\textsc{\textcolor{gray}{10x}26} &  \textbf{2.58} & 0.28 & 0.31 & 0.43 & 0.51 \\[1.2pt]
\bottomrule[1.5pt]
\end{tabular}}
\end{minipage}
\end{center}
\end{table}

\subsection{Model Size Variants}

In Table \ref{tab:model_size_variants}, we present four variants of the \lsm models we trained. The model sizes and naming conventions partially follow the tradition established by T5 \citep{raffel2020exploring}. Our results indicate that scaling the model beyond \lsm-B offers no additional improvements in either reconstruction loss or downstream task performance. Based on this insight all neural methods in Table~\ref{tab:task_results} employ a ViT-110M backbone.

\begin{table}[ht!]
\setlength{\tabcolsep}{4.5pt}
\caption{\small \textbf{Vision transformer size variants used in \lsm.} An \lsm-[size] model indicates a ViT-[size] backbone.}
\vspace{-1.5pt}
\label{tab:model_size_variants}
\small
\begin{center}
\begin{minipage}[t]{\textwidth}
\centering
\adjustbox{max width=\textwidth}{
\begin{tabular}{lcccccccc}
\toprule[1.5pt]
\multirow{2}{*}{\textbf{Model}} & \textbf{Encoder} & \textbf{Decoder} & \textbf{Encoder} & \textbf{Decoder} & \textbf{Encoder} & \textbf{Decoder} & \textbf{Total} & \multirow{2}{*}{\textbf{gFLOPs}} \\
 & \textbf{Blocks} & \textbf{Blocks} & \textbf{Dim} & \textbf{Dim} & \textbf{Heads} & \textbf{Heads} & \textbf{Params} & \\  
\midrule\midrule
\lsm-Tiny & 4 & 2  & 192 & 128 & 3 & 4 & 2M & 0.37 \\[1.2pt]
\lsm-Small & 8 & 2  & 256 & 192 & 4 & 4 & 7M & 1.28 \\[1.2pt]
\lsm-Base & 12 & 8  & 768 & 512 & 12 & 16 & 110M & 15.94 \\[1.2pt]
\lsm-Large & 24 & 8  & 1024 & 512 & 16 & 16 & 328M & 56.10 \\[1.2pt]
\bottomrule[1.5pt]
\end{tabular}}
\end{minipage}%
\end{center}
\end{table}

\section{Additional Results and Analysis}
\label{sec:additional_results}

\subsection{Results of Scaling Experiments for Generative Tasks}
\label{sec:appendix_gen_tasks_scaling}
\textbf{Generative Performance wrt. Data Scaling.} Table \ref{tab:appendix_gen_task_data_scaling} presents the full results for the generative tasks, evaluated across four model sizes and all data scales, including an experiment on the largest 40 million hour pretraining set. The \lsm Base model, trained on 6.6 million hours of data, achieved the best overall performance.

\textbf{Scaling Pretraining Data to 40 Million Hours.} As mentioned in Sections~\ref{sec:data_construction} and~\ref{sec:experiment_results}, we derive our dataset, used for presented scaling and downstream task results from 6.6M hours of data balanced across 160K people. To test the extremes of data scaling we also build a dataset comprising of 40M data hours by combining the 6.6M hours with an additional 33M hours of data from a 78569 subject subset of the total 160K subjects. However, as shown in Table \ref{tab:appendix_gen_task_data_scaling}, we observed that scaling benefits taper off when training the \lsm-Base model with this extended dataset. We believe this is due to two key factors: the structure of our dataset and the inherent limitations of the masking pretraining task, as discussed in Section~\ref{sec:limitation_and_future_work}. It is also possible that as the additional 33M hours are not evenly distributed across subjects that the careful balance of the 6.6M dataset is disturbed.

\subsection{Results of Scaling Experiments for Discriminative Tasks}
\label{sec:appendix_dis_tasks_scaling}

\begin{wrapfigure}{h!}{0.35\textwidth}
    \vspace{-5pt}  
    \centering
    \includegraphics[width=0.35\textwidth]{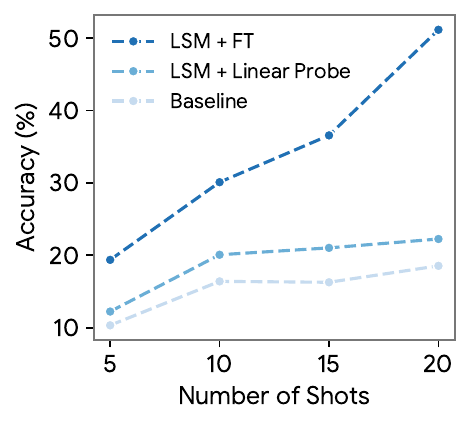}
    \caption{\textbf{Few shot learning.} Activity recognition results.}
    \label{fig:few_shot_ar}
    \vspace{-15pt}  
\end{wrapfigure}

\textbf{Discriminative Performance wrt. Data Scaling.} In Table \ref{tab:dis_task_data_scaling}, we demonstrate that scaling up the dataset significantly benefits downstream discriminative tasks, particularly in the fine-tuning stage. Furthermore, our pretrained \lsm model exhibits superior performance in label-efficient transfer learning, as shown in Table \ref{tab:few_shot_data}. Activity recognition few-shot results, as compared to a supervised baseline, are also visualized in Fig.~\ref{fig:few_shot_ar}. From the visualization it is clear that pretraining helps \lsm learn a strong representation of sensor data that enables more sample efficient performance on discriminative tasks.

\textbf{Convolutional Probe.} Following prior work~\citep{he2022masked} we explore an intermediary evaluation to linear probing and full-model fine-tuning. Specifically, we explore the learnable pooling of embeddings. This probe takes  patch-embeddings, produced by the encoder, and reshapes them to [num. patches $H$, num. patches $W$, embedding dimension], similar to the shape of the original patched sensor-image. This embedding is fed through two shallow convolutional layers and a linear head. We find that with less that 0.2\% of the trainable-parameters needed for full-model fine-tuning, we are able to achieve similar performance on exercise detection and activity recognition tasks. These results can be seen in Tables~\ref{tab:dis_task_data_scaling} and~\ref{tab:few_shot_data}.

\begin{table}[h!]
\small
\begin{center}
\begin{minipage}[t]{0.78\textwidth}
\centering
\caption{\small \textbf{Detailed Results of Generative Tasks.} Performance across Data and Model Sizes on Generative Tasks. Data Size is in hours.}
\vspace{-1.5pt}
\label{tab:appendix_gen_task_data_scaling}
\adjustbox{max width=\textwidth}{
\begin{tabular}{lcccc}
\toprule[1.5pt]
\multirow{3}{*}{\textbf{Data Size}} & \multirow{3}{*}{\textbf{Model Size}} & \multicolumn{3}{c}{\textbf{Task Error (MSE)}} \\
\cmidrule(lr){3-5}
 &  & \textbf{Random} & \textbf{Extrapolation} & \textbf{Interpolation} \\
 &  & \textbf{Imputation 80\%} & \textbf{60 min} & \textbf{60 min} \\
\midrule\midrule
\multirow{4}{*}{0.005 M}    & Tiny       & 0.50 & 0.71 & 0.53 \\
                         & Small      & 0.57 & 0.77 & 0.62 \\
                         & Base       & 0.67 & 0.80 & 0.68 \\
                         & Large      & 0.64 & 0.82 & 0.75 \\
\midrule
\multirow{4}{*}{0.05 M}   & Tiny       & 0.25 & 0.70 & 0.47 \\
                         & Small      & 0.29 & 0.58 & 0.36 \\
                         & Base       & 0.38 & 0.65 & 0.42 \\
                         & Large      & 0.38 & 0.65 & 0.43 \\
\midrule
\multirow{4}{*}{0.5 M}  & Tiny       & 0.22 & 0.62 & 0.42 \\
                         & Small      & 0.21 & 0.53 & 0.37 \\
                         & Base       & 0.22 & 0.48 & 0.28 \\
                         & Large      & 0.22 & 0.50 & 0.34 \\
\midrule
\multirow{4}{*}{3.8 M}  & Tiny       & 0.22 & 0.62 & 0.42 \\
                         & Small      & 0.21 & 0.49 & 0.36 \\
                         & Base       & 0.19 & 0.44 & 0.26 \\
                         & Large      & 0.21 & 0.64 & 0.46 \\
\midrule
\multirow{4}{*}{6.6 M} & Tiny       & 0.22 & 0.63 & 0.42 \\
                         & Small      & 0.21 & 0.49 & 0.35 \\
                         & Base       & 0.19 & 0.44 & 0.26 \\
                         & Large      & 0.20 & 0.54 & 0.40 \\
\midrule
40 M    & Base       & 0.19 & 0.45 & 0.27 \\
\toprule[1.5pt]
\end{tabular}}
\end{minipage}
\end{center}
\end{table}

\begin{table}[h!]
\setlength{\tabcolsep}{2.5pt}
\caption{\small \textbf{Data Scaling on Discriminative Tasks.}}
\vspace{-1.5pt}
\label{tab:dis_task_data_scaling}
\small
\begin{center}
\begin{minipage}[t]{0.78\textwidth}
\centering
\adjustbox{max width=\textwidth}{
\begin{tabular}{llcccc}
\toprule[1.5pt]
\multirow{2}{*}{\textbf{Data Size}} & \multirow{2}{*}{\textbf{Method}} & \multicolumn{2}{c}{\textbf{Exercise Detection}} & \multicolumn{2}{c}{\textbf{Activity Recognition}} \\
\cmidrule(r){3-4}\cmidrule(r){5-6}
& & \textbf{Accuracy} & \textbf{mAP} & \textbf{Accuracy} & \textbf{mAP} \\  
\midrule\midrule
\textsc{0.005 M} & \multirow{5}{*}{Linear Probe} & 60.6 & 49.8 & 35.1 & 17.5 \\[1.2pt]
\textsc{0.05 M} & & 67.3 & 61.0 & 39.6 & 23.4 \\[1.2pt]
\textsc{0.5 M} & & 84.5 & 78.8 & 47.1 & 24.7 \\[1.2pt]
\textsc{3.8 M} & & 88.0 & 85.0 & 47.6 & 25.3 \\[1.2pt]
\textsc{6.6 M} & & 84.7 & 89.0 & 49.4 & 24.6 \\[1.2pt]
\midrule
\textsc{0.005 M} & \multirow{5}{*}{Convolutional Probe} & 71.3 & 71.8 & 50.9 & 25.1 \\[1.2pt]
\textsc{0.05 M} & & 78.0 & 82.3 & 62.2 & 43.7 \\[1.2pt]
\textsc{0.5 M} & & 88.2 & 96.4 & 68.1 & 45.5 \\[1.2pt]
\textsc{3.8 M} & & 88.2 & 96.4 & 70.5 & 47.1 \\[1.2pt]
\textsc{6.6 M} & & 87.5 & 95.8 & 67.6 & 48.5 \\[1.2pt]
\midrule
\textsc{0.005 M} & \multirow{5}{*}{Fine Tune} & 68.3 & 58.9 & 51.5 & 30.0 \\[1.2pt]
\textsc{0.05 M} & & 73.8 & 77.0 & 64.0 & 48.0 \\[1.2pt]
\textsc{0.5 M} & & 84.9 & 93.7 & 68.8 & 50.0 \\[1.2pt]
\textsc{3.8 M} & & 87.5 & 96.4 & 64.2 & 48.7 \\[1.2pt]
\grayrow
\textsc{6.6 M} & & 90.3 & 97.0 & 68.5 & 51.4 \\[1.2pt]

\bottomrule[1.5pt]
\end{tabular}}
\end{minipage}%
\end{center}
\end{table}

\begin{table}[h!]
\setlength{\tabcolsep}{2.5pt}
\caption{\small \textbf{Few-Shot Performance on Discriminative Tasks.}}
\vspace{-1.5pt}
\label{tab:few_shot_data}
\small
\begin{center}
\begin{minipage}[t]{0.78\textwidth}
\centering
\adjustbox{max width=\textwidth}{
\begin{tabular}{llccccc}
\toprule[1.5pt]
\multirow{2}{*}{\textbf{Samples per Class}} & \multirow{2}{*}{\textbf{Method}} & \multicolumn{2}{c}{\textbf{Exercise Detection}} & \multicolumn{2}{c}{\textbf{Activity Recognition}} \\
\cmidrule(r){3-4}\cmidrule(r){5-6}
& & \textbf{Accuracy} & \textbf{mAP} & \textbf{Accuracy} & \textbf{mAP} \\  
\midrule\midrule
\textsc{5} & \multirow{4}{*}{Linear Probe} & 51.3 & 48.0 & 12.2 & 17.5 \\[1.2pt]
\textsc{10} && 58.3 & 57.1 & 20.1 & 18.4 \\[1.2pt]
\textsc{15} && 65.4 & 68.8 & 21.0 & 18.7 \\[1.2pt]
\textsc{20} && 65.1 & 69.8 & 22.3 & 18.8 \\[1.2pt]
\midrule
\textsc{5} & \multirow{4}{*}{Convolutional Probe} & 40.5 & 43.8 & 20.6 & 24.7 \\[1.2pt]
\textsc{10} && 63.2 & 59.4 & 27.9 & 26.7 \\[1.2pt]
\textsc{15} && 57.3 & 60.8 & 27.9 & 26.7 \\[1.2pt]
\textsc{20} && 67.0 & 56.9 & 36.9 & 25.3 \\[1.2pt]
\midrule
\textsc{5} & \multirow{4}{*}{Fine Tune} & 54.7 & 56.8 & 19.4 & 21.5 \\[1.2pt]
\textsc{10} && 65.8 & 65.1 & 30.1 & 22.7 \\[1.2pt]
\textsc{15} && 71.1 & 73.1 & 36.6 & 24.8 \\[1.2pt]
\textsc{20} && 65.6 & 67.1 & 51.2 & 33.2 \\[1.2pt]
\midrule
\textsc{5} & \multirow{4}{*}{Supervised} & 43.1 & 52.9 & 10.3 & 14.5 \\[1.2pt]
\textsc{10} && 49.3 & 46.0 & 16.4 & 14.6 \\[1.2pt]
\textsc{15} && 49.6 & 50.6 & 16.3 & 14.4 \ \\[1.2pt]
\textsc{20} && 48.2 & 45.8 & 18.5 & 23.0 \\[1.2pt]

\bottomrule[1.5pt]
\end{tabular}}
\end{minipage}%
\end{center}
\end{table}

\subsection{Classification Confusion Matrices}
\label{appendix-sec:confusion_matrices}
\begin{figure*}[h!]
    \centering
    \includegraphics[width=0.70\textwidth]{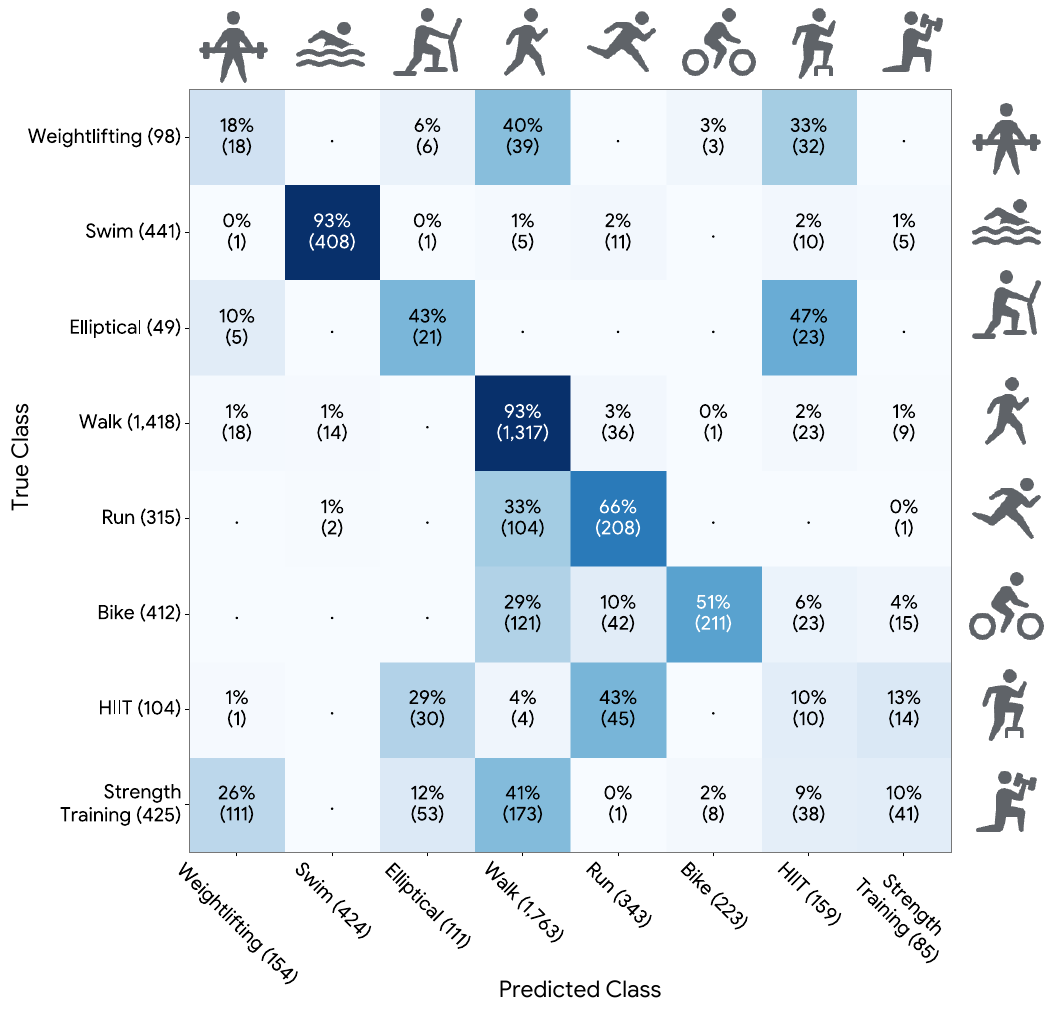}
    \caption{\textbf{Activity recognition confusion matrix.} Results for the full-model fined-tuned \lsm Masked Auto-Encoder.}
    \label{fig:confusion_matrix} 
\end{figure*}

Fig.~\ref{fig:confusion_matrix} presents the complete confusion matrix for our activity recognition task from the full-model fine-tuned \lsm-B model. Note that many classes get mistaken for \textit{Walk}. This is likely as there are significant periods of walking in the 5-hour inputs, even if the activity is labeled otherwise.

\subsection{Feature Embeddings}
\label{appendix:feature_embeddings}

\begin{figure*}[h!]
    \centering
    \includegraphics[width=\textwidth]{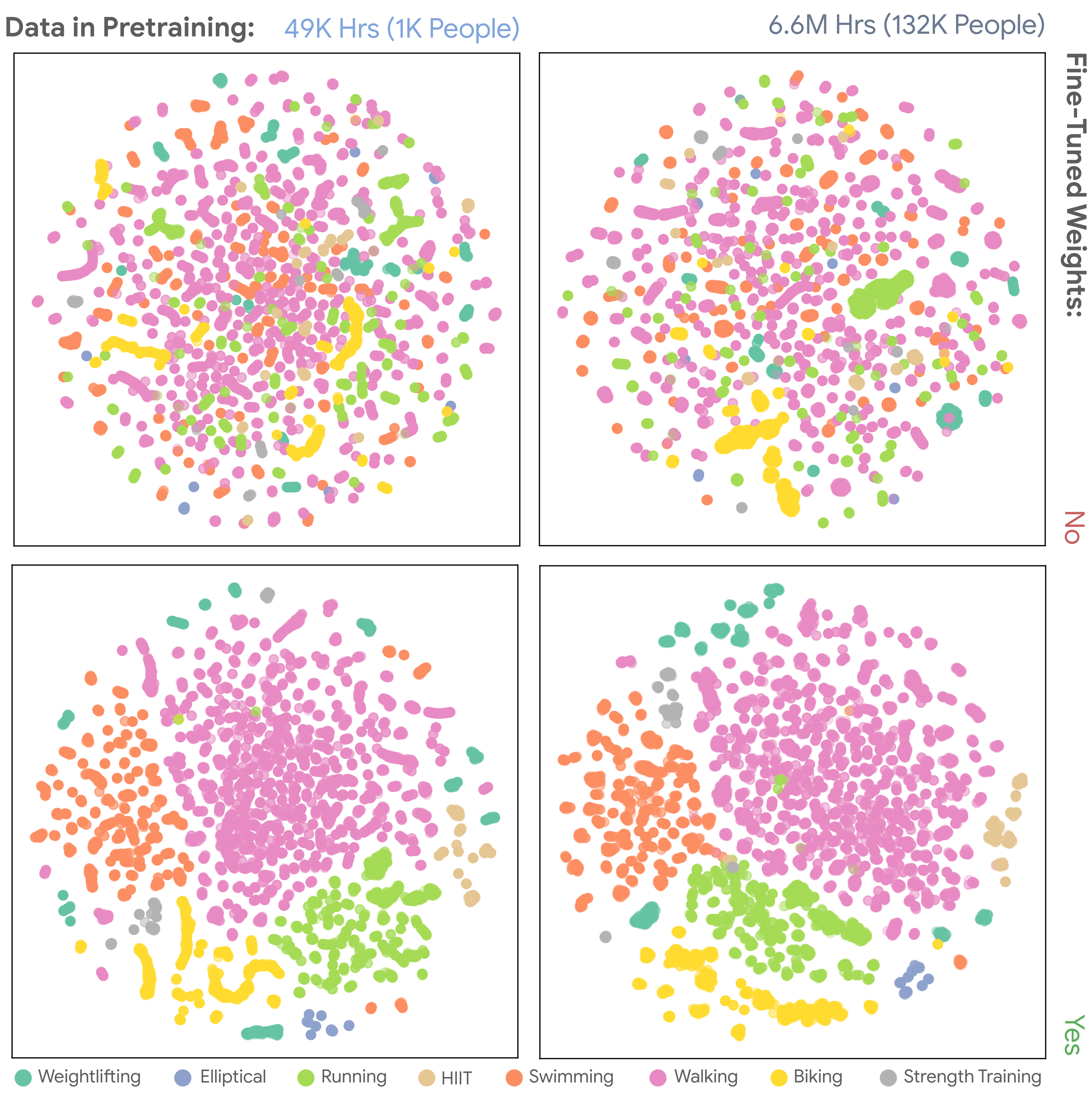}
    \caption{\textbf{t-SNE Embeddings for Pretraining and Fine-tuned Models Labeled by Activity.} t-distributed Stochastic Neighbor Embedding (t-SNE) plots showing that there are differences (albeit subtle) between pretrained embeddings using data from almost 50k and 6.6 hours.}
    \label{fig:activity_tsne} 
\end{figure*}

\begin{figure*}[h!]
    \centering
    \includegraphics[width=\textwidth]{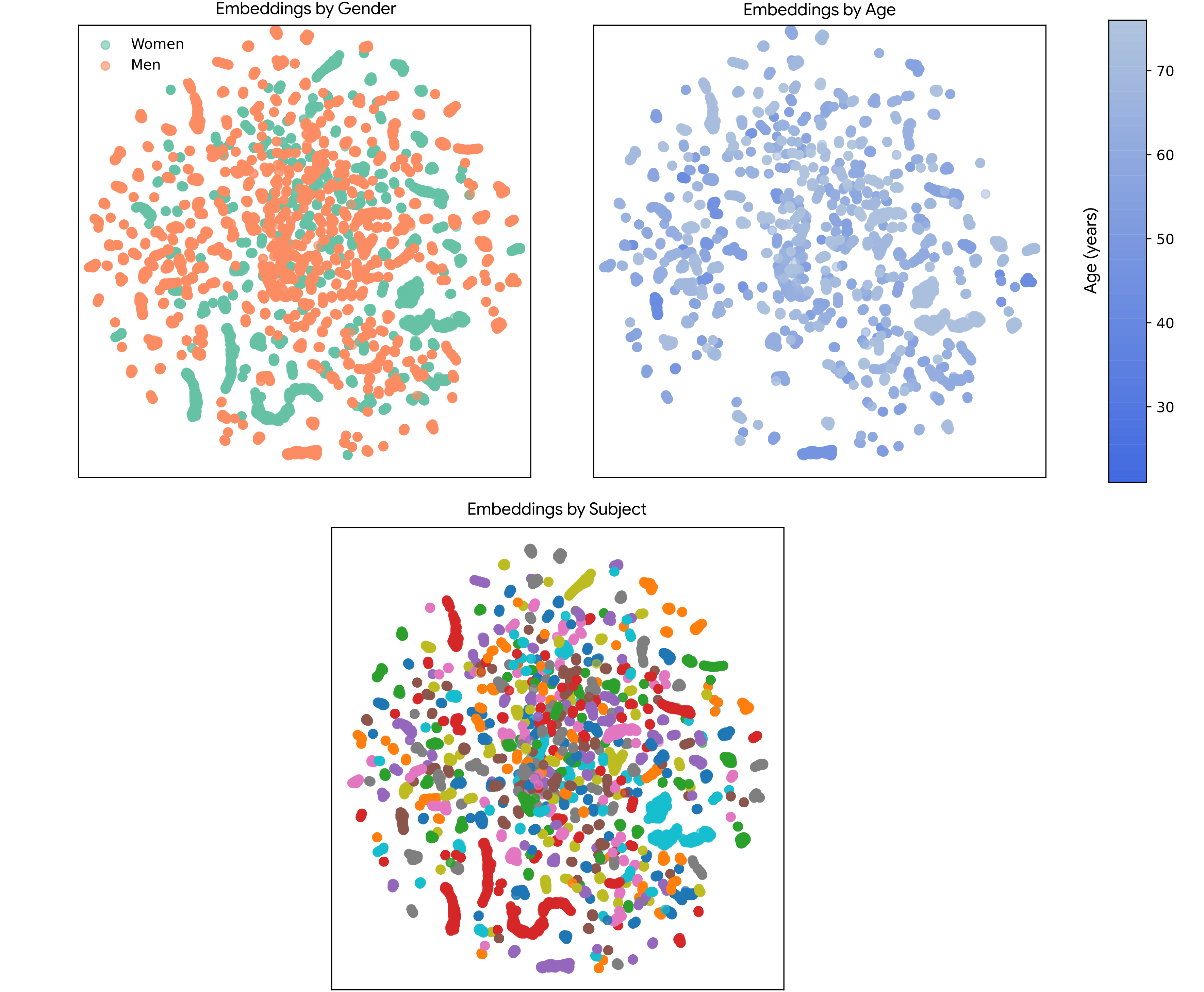}
    \caption{\textbf{t-SNE Embeddings Labeled by Gender, Age and Subject.} t-distributed Stochastic Neighbor Embedding (t-SNE) plots showing that the learned embeddings do capture subject specific information (and therefore also exhibit some subtle gender and age clusters). Age was not available for all subjects.}
    \label{fig:demographic_tsne} 
\end{figure*}

We present t-distributed Stochastic Neighbor Embedding (t-SNE) plots. In Fig.~\ref{fig:activity_tsne} illustrates that scaling pretraining data results in noticeable, albeit subtle, improvements of clustering across activities in the learned representation. We also find that fine-tuning the model is critical to effectively discriminate between activities.
In Fig.~\ref{fig:demographic_tsne} we see that the learned representation does embed some subject dependencies. This can be attributed to variance in the physiology and activity definitions for individuals (e.g., a hard run may look very different for two different people).

\subsection{Signal Correlations}
\label{appendix-sec:signal_correlations}

\begin{figure}[h!]
    \centering
    \includegraphics[width=\textwidth]{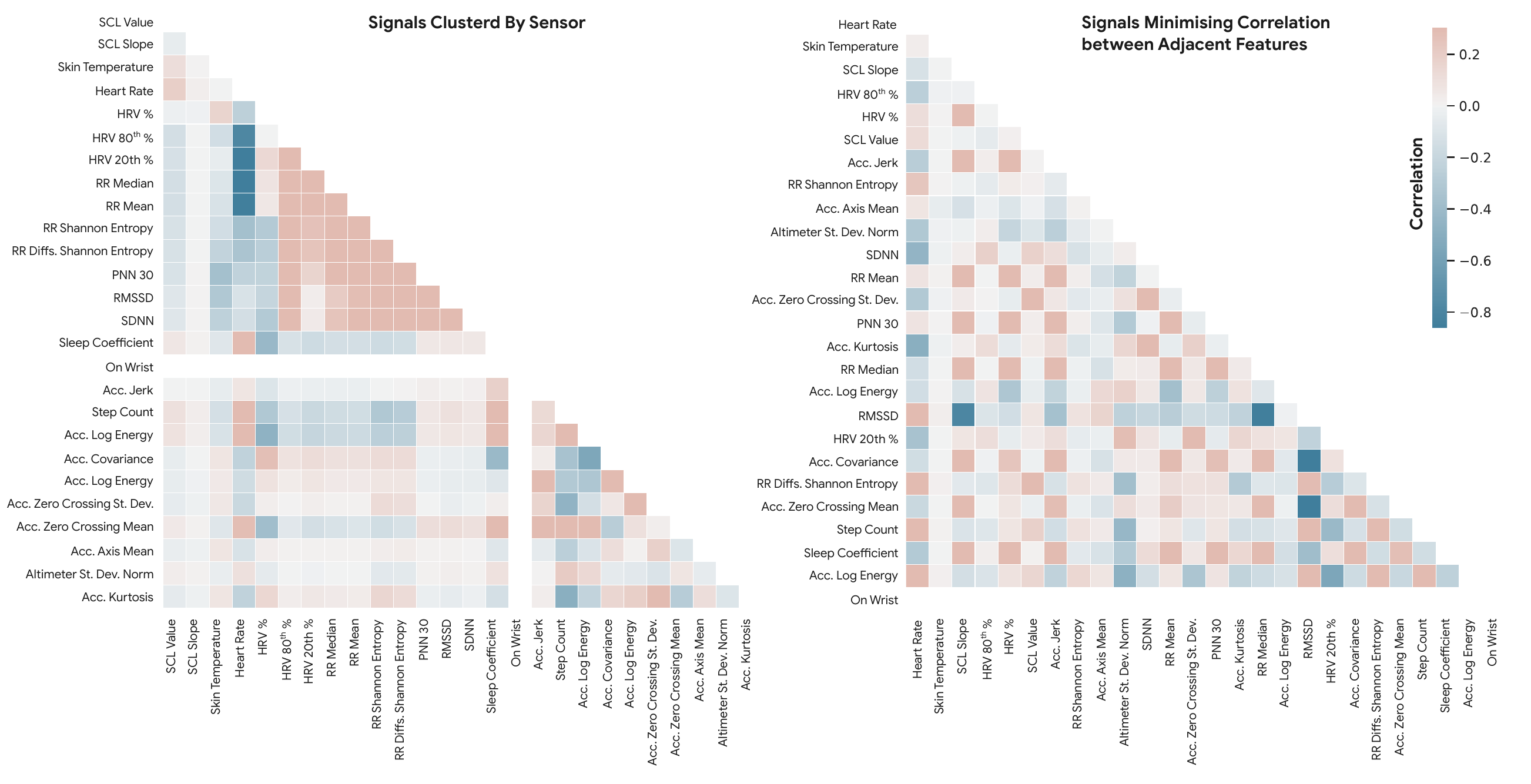}
    \caption{\textbf{Sensor Signal Diagonal Correlation Matrix.} The pair-wise correlation between the 26 sensor features based on our training set.}
    \label{fig:feature_correlations} 
\end{figure}

The 26 signals used as input to our model come from four sensors (accelerometer, PPG, temperature, altimeter). As a result, some signals are more correlated with certain ones than with others.. A signal diagonal correlation matrix was calculated to show the pairwise correlations between signals. Fig.~\ref{fig:feature_correlations} shows the correlation matrix for signals clustered by sensor and for signals ordered to minimize the absolute correlation coefficient between adjacent features.

\subsection{Examples of Reconstructions}
\label{appendix-sec:reconstructions}

\begin{figure}[h!]%
\centering
\includegraphics[width=0.85\textwidth]{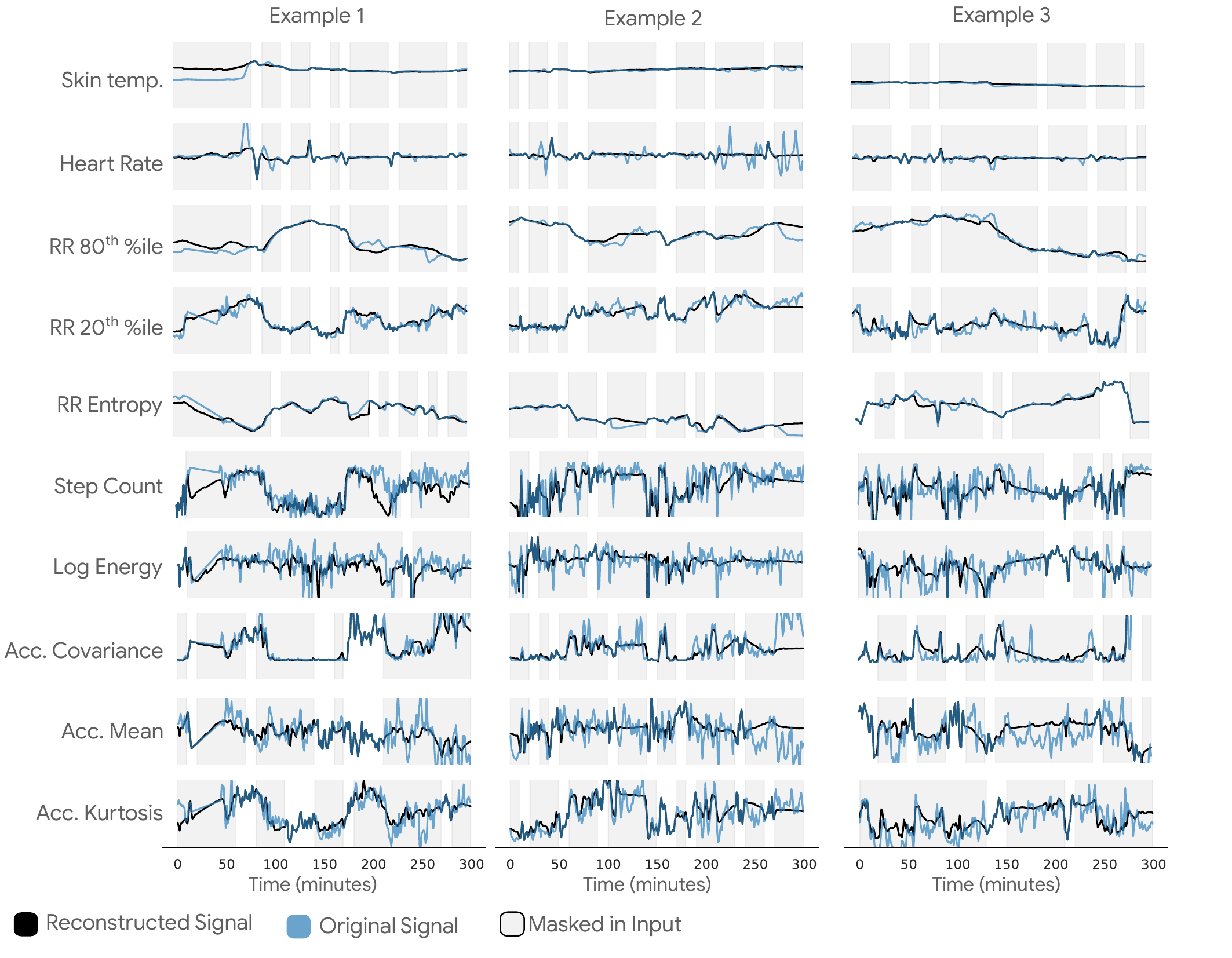}
\caption{\textbf{Example of Signal Reconstructions.} Comparison between the ground-truth (blue) and reconstruction (black) for a 5-hour sample. Gray regions were masked in the input. 80\% Random Masking (Patch Size 10 mins x 5 sensors). Note: model outputs are only shown for the masked regions in the reconstructions.}%
\label{fig:reconstruction_examples}
\end{figure}

\begin{figure}[h!]%
\centering
\includegraphics[width=\textwidth]{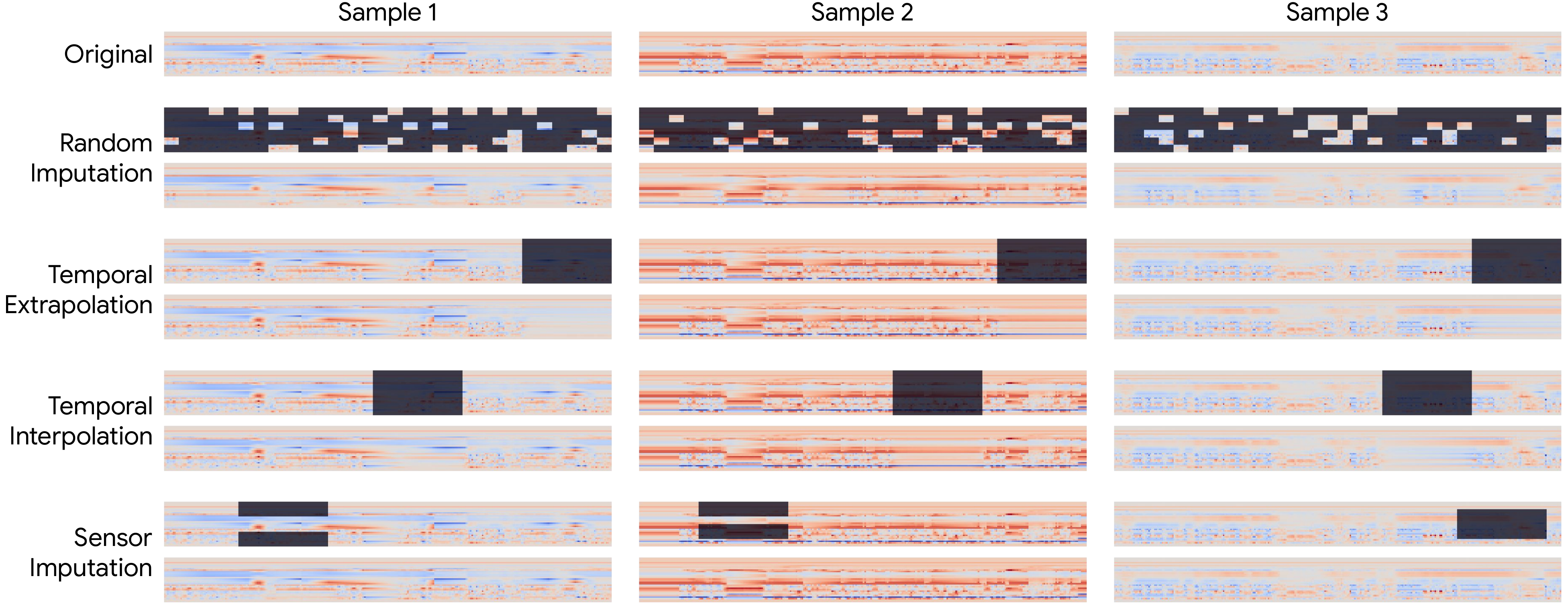}
\caption{\textbf{Examples of Signal Reconstructions Across Generative Down Stream Tasks.} The top row of each sample shows the original sensor signal image. Subsequent row-pairs plot the masked input followed by the model reconstruction below. All reconstruction come from \lsm-Base based \lsm employing a 10x5 patch size and pretrained with 80\% random masking. Note: model outputs are only shown for the masked patches in the reconstructions.}%
\label{fig:task_reconstruction_examples}
\end{figure}

\begin{figure}[h!]%
\centering
\includegraphics[width=\textwidth]{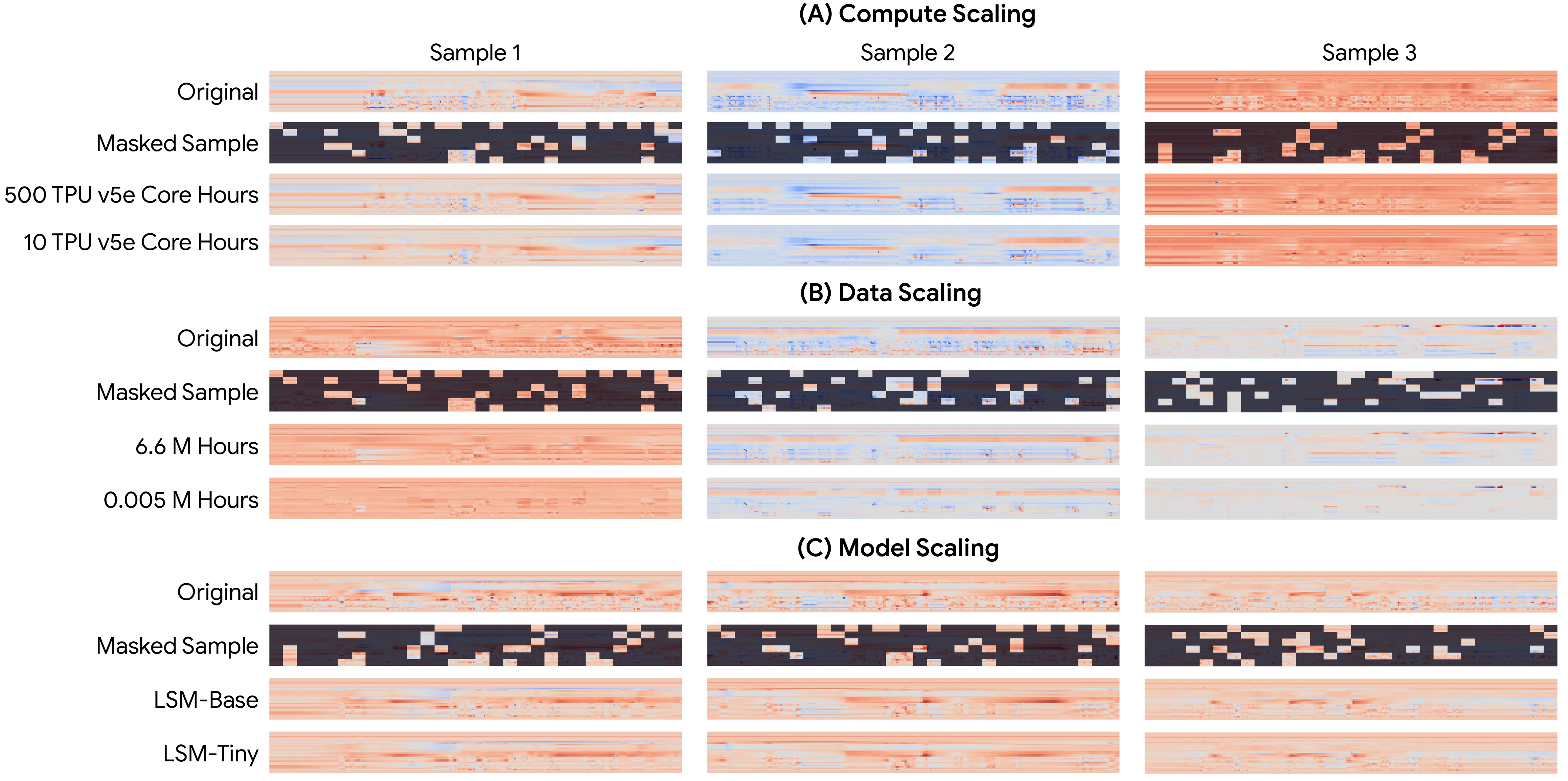}
\caption{\textbf{Examples of Signal Reconstructions with Respect to Scaling.} These plots illustrate the (often visually subtle) affect of \textbf{A} compute,  \textbf{B} data, and \textbf{C} model scaling for sensor models. Note: model outputs are only shown for the masked patches in the reconstructions.}%
\label{fig:scaling_reconstruction_examples}
\end{figure}

A qualitative example of ground-truth signals and corresponding reconstructions are shown in Fig.~\ref{fig:reconstruction_examples}.  The gray regions are sections that were masked in the input. Additional sensor-image level reconstructions, across generative down-stream tasks (eg. imputation, extrapolation, interpolation) can be seen in Fig.~\ref{fig:task_reconstruction_examples}. Examples of the often visually subtle affects of scaling on reconstruction can be seen in Fig. ~\ref{fig:scaling_reconstruction_examples}.

\section{Details of Training and Hyperparameters}
\label{sec:appendix_hyperparas}
\textbf{Hyperparameters}. This section provides details about the pretraining and fine-tuning of \lsm and other baseline methods. The pretraining hyperparameters, detailed in Table \ref{tab:lsm_pretraining_settings}, were chosen with hyperparameter sweeps. In Table \ref{tab:hyper_params_fine_tune_linear_probe}, we include hyperparameters for linear probe and fine-tuning. The hyperparameters for supervised baseline training are detailed in Table \ref{tab:hyper_params_supervised_training}. Note that hyperparameters used for the few-shot experiments found in Table~\ref{tab:few_shot_data} are similar to those found in Tables~\ref{tab:hyper_params_fine_tune_linear_probe} and~\ref{tab:hyper_params_supervised_training} with slight changes in learning rate.

\textbf{Training Augmentations}. Traditional image augmentations are not always valid when applied to sensor-images. For example, random crop and resize, often applied in contrastive pretraining are invalid for sensor images, as a random crop may remove a subset of senor signals. Thus, we define a subset of augmentations valid for sensor-images. These are random \textit{Flip}: a flip along the temporal axis; \textit{Stretch}: a stretch along the temporal axis of and subsequent crop of back to original time length; and \textit{Noise}: the addition of Gaussian noise.

\begin{table}[h!]
    \centering
    \caption{\small\textbf{Hyperparameters for pretraining} with MAE \citep{he2022masked}, MSN \citep{assran2022masked}, DINO \citep{caron2021emerging} and SimCLR \citep{chen2020simple}. A solitary row value indicates that the value was used for all methods.}
    \label{tab:lsm_pretraining_settings}
    \begin{tabular}{lcccc}
    \toprule[1.5pt]
        \textbf{Configuration} & MAE & MSN & DINO & SimCLR \\
        \midrule\midrule
        Training Steps & \multicolumn{4}{c}{50000} \\
        Warmup Steps& \multicolumn{4}{c}{2500} \\
        Optimizer                  & \multicolumn{4}{c}{AdamW \citep{loshchilov2017decoupled}} \\
        Opt. momentum [$\beta_1, \beta_2$]              & [0.9, 0.95] & [0.9, 0.99] & [0.9, 0.99] & [0.9, 0.99] \\
        Base learning rate & 0.005 & 0.001 & 0.004 & 0.001 \\
        Batch size                 & \multicolumn{4}{c}{4096} \\
        Weight decay               & \multicolumn{4}{c}{0.0001} \\
        Gradient clipping          & 1.0 & 3.0 & 3.0 & 3.0 \\
        Dropout                    & \multicolumn{4}{c}{0.0} \\
        Learning rate schedule     & \multicolumn{4}{c}{Linear Warmup \& Cosine Decay} \\
        Loss Function              & \multicolumn{4}{c}{Mean Squared Error} \\
        Data resolution           & \multicolumn{4}{c}{ 26 (signal)$\times$300(minute)} \\
        Augmentation               & \multicolumn{4}{c}{Flip, Stretch, Noise} \\
        
    \toprule[1.5pt]
    \end{tabular}
\end{table}

\begin{table}[h!]
    \centering
    \caption{\small\textbf{Hyperparameters for Linear Probing and Fine-Tuning on Discriminative Tasks} detailed in Section \ref{sec: discriminative_task}. A solitary row value indicates that it was used for all methods. LP=Linear Probe. FT=Fine-Tune (full model).}
    \label{tab:hyper_params_fine_tune_linear_probe}
    \begin{tabular}{lcccc}
    \toprule[1.5pt]
        \textbf{Task} & \multicolumn{2}{c}{Exercise Detection} & \multicolumn{2}{c}{Activity Recognition} \\
        \textbf{Configuration} & LP & FT & LP & FT \\
        \midrule\midrule
        Training Steps & 400 & 400 &  300 & 300 \\
        Warmup Step Percent & 20 & 20 & 15 & 15 \\
        Optimizer & \multicolumn{4}{c}{AdamW \citep{loshchilov2017decoupled}} \\
        Opt. momentum [$\beta_1, \beta_2$]              & \multicolumn{4}{c}{[0.9, 0.95]} \\
        Base learning rate & 0.5 & 0.00005 & 0.5 & 0.00005 \\
        Batch size                 & \multicolumn{4}{c}{128} \\
        Weight decay               & \multicolumn{4}{c}{0.0001} \\
        Gradient clipping          & \multicolumn{4}{c}{1.0} \\
        Dropout                    & \multicolumn{4}{c}{0.3} \\
        Learning rate schedule     & \multicolumn{4}{c}{Linear Warmup \& Cosine Decay} \\
        Loss Function              & \multicolumn{4}{c}{Balanced Softmax Loss \citep{Ren2020balms}} \\
        Data resolution           & \multicolumn{4}{c}{26 (signal)$\times$300(minute)} \\
        Augmentation               & \multicolumn{4}{c}{Noise} \\
    \toprule[1.5pt]
    \end{tabular}
\end{table}

\begin{table}[h!]
    \centering
    \caption{\small\textbf{Hyperparameters for Supervised Training on Discriminative Tasks} A solitary row value indicates that it was used for all methods.}
    \label{tab:hyper_params_supervised_training}
    \begin{tabular}{lcc}
    \toprule[1.5pt]
        \textbf{Configuration} & Exercise Detection & Activity Recognition \\
        \midrule\midrule
        Training Steps & 400 & 300 \\
        Warmup Steps& 20 & 15 \\
        Optimizer                  & \multicolumn{2}{c}{AdamW \citep{loshchilov2017decoupled}} \\
        Opt. momentum [$\beta_1, \beta_2$] & \multicolumn{2}{c}{[0.9, 0.95]} \\
        Base learning rate         & 0.0001 & 0.0005 \\
        Batch size                 & \multicolumn{2}{c}{128} \\
        Weight decay               & \multicolumn{2}{c}{0.0001} \\
        Gradient clipping          & \multicolumn{2}{c}{1.0} \\
        Dropout                    & \multicolumn{2}{c}{0.0} \\
        Learning rate schedule     & \multicolumn{2}{c}{Linear Warmup \& Cosine Decay} \\
        Loss Function              & \multicolumn{2}{c}{Balanced Softmax Loss \citep{Ren2020balms}} \\
        Data resolution           & \multicolumn{2}{c}{26 (signal)$\times$300(minute)} \\
        Augmentation               & \multicolumn{2}{c}{Noise} \\
    \toprule[1.5pt]
    \end{tabular}
\end{table}

\section{Description of Pretraining and Baseline Methods}

\subsection{Pretraining Methods}

There are two main approaches to pretraining, one based on contrastive learning and the other based on the reconstruction or prediction of input features. At a high-level contrastive methods where representations are learned for different views of the same training example (\textit{positives}), and dissimilar embeddings for different training examples (\textit{negatives}). However, there are challenges or drawbacks to this approach. First, in the sensor domain it can be non-trivial to create augmentations that do not alter the meaning (label) of the sample. For example, does stretching data for someone \emph{running} mean that it more closely resembles the data when they \emph{walk}? Second, generative capabilities are attractive as imputing missing data and forecasting signals into the future are useful in and or themselves. As such, purely contrastive set-up has limitations and a pretraining task based on the reconstruction of masked input tokens is attractive. A masked autoencoder is one example of such an approach that is effective at scalable learning of representations~\citep{he2022masked}. Below we describe the pretraining methods used for \lsm and our baselines.

\textbf{Masked Auto Encoder (MAE) \citep{he2022masked}.}  
MAE is a self-supervised learning method where the input data is randomly masked, and the model is trained to reconstruct the missing parts. It operates on the principle that forcing the model to predict missing information helps it learn meaningful representations. MAE has shown strong performance in various vision and signal tasks, particularly in cases where large-scale unlabeled data is available.

\textbf{SimCLR \citep{chen2020simple}.}  
SimCLR is a contrastive learning framework that learns representations by maximizing agreement between different augmented views of the same data sample. The method uses a contrastive loss, which encourages the model to pull together similar views of the same sample while pushing apart views of different samples. SimCLR has been widely used in both vision and sensor data for representation learning without requiring labeled data.

\textbf{Masked Siamese Network (MSN) \citep{assran2022masked}.} 
MSN combines the benefits of invariance-based pretraining with mask denoising. MSN operates by matching the representation of an image view with randomly masked patches to the representation of the corresponding unmasked image. This pretraining strategy leverages Vision Transformers by processing only the unmasked patches, significantly enhancing scalability. The framework enables the generation of semantically rich representations, which perform competitively in low-shot image classification tasks.

\textbf{DINO \citep{caron2021emerging}.}  
DINO is a self-distillation method that trains the model using knowledge distillation, without the need for labeled data. It leverages a teacher-student network architecture, where the teacher generates target representations for the student to learn from. DINO has demonstrated success in generating robust representations that can be transferred to various downstream tasks.

\subsection{Generative Baselines}

We define a number of baselines for our generative tasks. Similar methods are common-place in the image domain (often used for up-sampling)~\citep{han2013comparison}, and the Internet of Things (IoT) sensor domain (often for imputing corrupted and/or missing data)~\citep{adhikari2022comprehensive}.

\textbf{Mean Fill.}  
Mean Fill is a simple baseline for generative tasks, where the missing values for a sensor stream are replaced by the mean value of the sensor data present in a given sample. Though naive, this method provides a reasonable estimate in certain contexts where missing values are randomly distributed.

\textbf{Nearest Neighbor Fill.}  
Nearest Neighbor Fill imputes missing data by using the value of the nearest observed neighbor for a given feature along the temporal axis. In the absence of a past and future neighbors this method mirrors back/forward fill. This method works well when there is a high degree of local similarity in the data.

\textbf{Linear Interpolation.}  
Linear Interpolation fills missing values by interpolating linearly between known values along the temporal dimension.  In the absence of a past and future neighbors this method mirrors back/forward fill. This baseline is often used in time-series and spatial data, where the assumption is that changes between data points occur in a smooth, continuous manner. 

For all generative baseline methods, in the rare cases where the sensor feature is completely missing, the feature values are replaced with zeros. This remains a valid strategy as all features are z-score normalized and centered around zero.

\subsection{Classification Baselines}

\textbf{Vision Transformer (ViT) \citep{dosovitskiy2020image}.}  
The Vision Transformer (ViT) is a transformer-based architecture that treats image patches or signal segments as input tokens, similar to how transformers handle sequences in natural language processing. ViT has shown competitive performance across various classification tasks, especially when trained with large amounts of data, and serves as a strong baseline in both vision and sensor classification tasks.

\section{Additional Details of Dataset}
\label{sec:dataset_additional_details}

In Table~\ref{tab:features} we detail the 26 derived sensor signal features leveraged by out method.

\begin{table}[ht!]
\small
    \centering
    \caption{\textbf{Sensor Feature Definitions.} Names, units and definitions of the 26 Accelerometer, PPG, skin conductance and altimeter features we use.}
    \begin{tabular}{rlp{7cm}}
    \toprule
        \textbf{Feature}   & \textbf{Unit} & \textbf{Definition} \\
        \midrule
        \grayrow
        \multicolumn{1}{l}{\EDACircle \textbf{Skin Conductance}} & & \\
        Skin Conductance Value & $\mu$Siemens & Center of linear tonic SCL value fit. \\ 
        Skin Conductance Slope & $\mu$S/Min & Intraminute slope of SCL values. \\ 
        \grayrow
        \multicolumn{1}{l}{\TempCircle \textbf{Skin Temperature}} & & \\
        Skin Temperature Value & \degree C & Mean skin temperature. \\ 
        \grayrow
        \multicolumn{1}{l}{\PPGCircle \textbf{Photoplethysmography}} & & \\
        Heart Rate & Beats/Min & Mean of instantaneous heart rate. \\ 
        RR Percent Valid & \% & \% of 5-minute window with valid RR intervals.  \\ 
        RR 80$^{th}$ Percentile & Msec & 80$^{th}$ percentile of 5-minute window of RR ints. \\ 
        RR 20$^{th}$ Percentile & Msec & 20$^{th}$ percentile of RR ints. \\ 
        RR Median & Msec & Median RR interval. \\ 
        RR Mean & Msec & Mean RR interval.\\ 
        Shannon Ent. RR & Nats & Shannon entropy of the RR intervals.$^{**}$ \\ 
        Shannon Ent. RR Diffs  & Nats & Shannon entropy of the RR interval differences.$^{**}$ \\ %
        PNN30 & \% & \% of successive RR ints. that change by $>$ 30ms. \\ 
        RMSSD & Msec & Root mean squared st. dev. of RR ints. \\ 
        SDNN & Msec & Standard deviation of RR intervals. \\ 
        On Wrist & Boolean & If optical-sensor off-wrist within a 30-second window, then false. \\ 
        \grayrow
        \multicolumn{1}{l}{\AccelerometerCircle \textbf{Accelerometer}} & & \\
        Jerk Autocorrelation Ratio & a.u. & Ratio of lag=1 autocorrelation to energy in 1st 3-axis principal component.\\ 
        Step Count & Steps & Number of steps. \\ 
        Log Energy & a.u. & Log of sum of 3-axis root mean squared magnitude. \\ 
        Covariance Condition & a.u. & Estimate of condition number for 3-axis covariance matrix. \\ 
        Log Energy Ratio & a.u. & Log of ratio of sum of energy in 1st 3-axis principal component over energy of 3-axis root mean squared magnitude.  \\ 
        Zero Crossing St.Dev. & Seconds & Standard deviation of time between zero crossing of 1st 3-axis principal component. \\ 
        Zero Crossing Average & Seconds & Mean of time between zero crossing of 1st 3-axis principal component.\\ 
        Robust Arm-Tilt & a.u. & Log of mean square root of squared X \& Z axes.\\ 
        Kurtosis & a.u. & Kurtosis of 3-axis root mean squared magnitude.\\
        Sleep Coefficient & a.u. & Sum of 3-axis max-min range, binned into 16 log-scaled bins.\\
        \grayrow  
        \multicolumn{1}{l}{\AltCircle
        \textbf{Altimeter}} & & \\
        Altimeter St.Dev. Norm & Hectopascals & Standard deviation of altimeter readings.\\ 
        \bottomrule
        \end{tabular}
    \label{tab:features}
\end{table}

\section{Code Acknowledgements}

We build our methods upon the \textit{Scenic} project~\citep{dehghani2021scenic}, an open source codebase for vision tasks implemented in JAX with Flax. \textit{Scenic} provides rich infrastructure for attention-base vision models and common vision baselines. The project page can be found here: \href{https://github.com/google-research/scenic}{github.com/google-research/scenic}.

\end{document}